\documentclass[10pt,twocolumn,letterpaper]{article}
\usepackage{iccv}
\usepackage{times}
\usepackage{epsfig}
\usepackage{graphicx}
\usepackage{amsmath}
\usepackage{amssymb}

\usepackage{url}
\usepackage{amsfonts}  
\usepackage{pifont} 
\usepackage{nicefrac}  
\usepackage{microtype} 
\usepackage{lipsum}
\usepackage{fancyhdr}  
\usepackage{booktabs} % professional-quality tables
\usepackage{multirow}
\usepackage{makecell}
\usepackage{environ}
\usepackage{comment}
\usepackage{floatrow}
\usepackage[accsupp]{axessibility}
\usepackage{adjustbox}
\usepackage{mathtools,kantlipsum}
\usepackage{caption}
\usepackage{subcaption}

\usepackage[pagebackref=true,breaklinks=true,letterpaper=true,colorlinks,bookmarks=false]{hyperref}
\usepackage[capitalize]{cleveref}

\iccvfinalcopy % *** Uncomment this line for the final submission

\ificcvfinal\fi

\newcommand{\mainBoost}{4.3\%}
\newcommand{\scanReferMainBoost}{5.0\%}

\newcommand{\datasetName}{ScanEnts3D}
\newcommand{\datasetSuffix}{ScanEnts}
\newcommand{\nr}[1]{Nr3D}
\newcommand{\sr}[1]{Sr3D}
\newcommand{\scanRefer}[1]{ScanRefer}
\newcommand{\scanReferTotalUtterances}{46,173}
\newcommand{\numberOfAnnotatedUtterances}[1]{37,842}
\newcommand{\numberOfEntitiesScanRefer}[1]{182k}
\newcommand{\numberOfEntities}[1]{96k}
\newcommand{\AverageLengthOfObjectsPerEntityScanRefer}[1]{1.33}
\newcommand{\AverageLengthOfEntitiesPerUtterance}[1]{2.55}
\newcommand{\AverageLengthOfObjectsPerEntity}[1]{1.32}
\newcommand{\AverageLengthOfWordsPerEntity}[1]{1.11}

\newfloatcommand{capbtabbox}{table}[][\FBwidth]
\NewEnviron{myresizeenv}{\resizebox{\linewidth}{!}{\BODY}}

\newcommand{\cmark}{\ding{51}}%
\newcommand{\xmark}{\ding{55}}%

\crefname{section}{Sec.}{Secs.}
\Crefname{section}{Section}{Sections}
\Crefname{table}{Table}{Tables}
\crefname{table}{Tab.}{Tabs.}

\newcommand{\webpage}{\url{https://scanents3d.github.io/}}
\begin{document}

%%%%%%%%%%%%%%%%%%%%%%%%%%%%%%%%%%%%%%%%%%%%%%%%%%
% Title & Authors
%%%%%%%%%%%%%%%%%%%%%%%%%%%%%%%%%%%%%%%%%%%%%%%%%%
\title{ScanEnts3D: Exploiting Phrase-to-3D-Object Correspondences for Improved Visio-Linguistic Models in 3D Scenes}

\author{Ahmed Abdelreheem$^{1,2}$, Kyle Olszewski$^{2}$, Hsin-Ying Lee$^{2}$, Peter Wonka$^{1}$, Panos Achlioptas$^{2}$\\
$^{1}$ King Abdullah University of Science and Technology (KAUST) \\
$^{2}$ Snap Inc.\\
\tt \small \{asamirh.95,olszewski.kyle,james371507,pwonka,pachlioptas\}@gmail.com
}

\maketitle

%%%%%%%%% ABSTRACT
\begin{abstract}

The two popular datasets \scanRefer{} \cite{zhenyu2019scanrefer} and ReferIt3D \cite{achlioptas2020referit_3d} connect natural language to real-world 3D scenes.
In this paper, we curate a complementary dataset extending both the aforementioned ones. We associate \textit{all} objects mentioned in a referential sentence with their underlying instances inside a 3D scene. In contrast, previous work did this only for a single object per sentence. 
Our \textbf{Scan} \textbf{Ent}itie\textbf{s} in \textbf{3D} ({\textit{\textbf{\datasetName{}}}}) dataset provides explicit correspondences between 369k objects across 84k referential sentences,  covering 705 real-world scenes. 
We propose novel architecture modifications and losses that enable learning from this new type of data and improve the performance for both \textit{neural listening} and \textit{language generation}. For neural listening, we improve the SoTA in \textit{both} the \nr{} and \scanRefer{} benchmarks by \textbf{\mainBoost{}} and \textbf{5.0\%}, respectively. For \textit{language generation}, we improve the SoTA by \textbf{13.2} CIDEr points on the \nr{} benchmark.
For both of these tasks, the new type of data is only used to improve training, but no additional annotations are required at inference time. The project's webpage is \webpage{}.

\end{abstract}

%%%%%%%%% BODY TEXT
\section{Introduction}
\begin{center}
  \textit{``The limits of my language mean the limits of my world.''}

  \textit{— Ludwig Wittgenstein.}
\end{center}
\begin{figure}[!htb]
     
    \subfloat[\centering Examples of \nr{}]{{\includegraphics[width=0.98\linewidth]{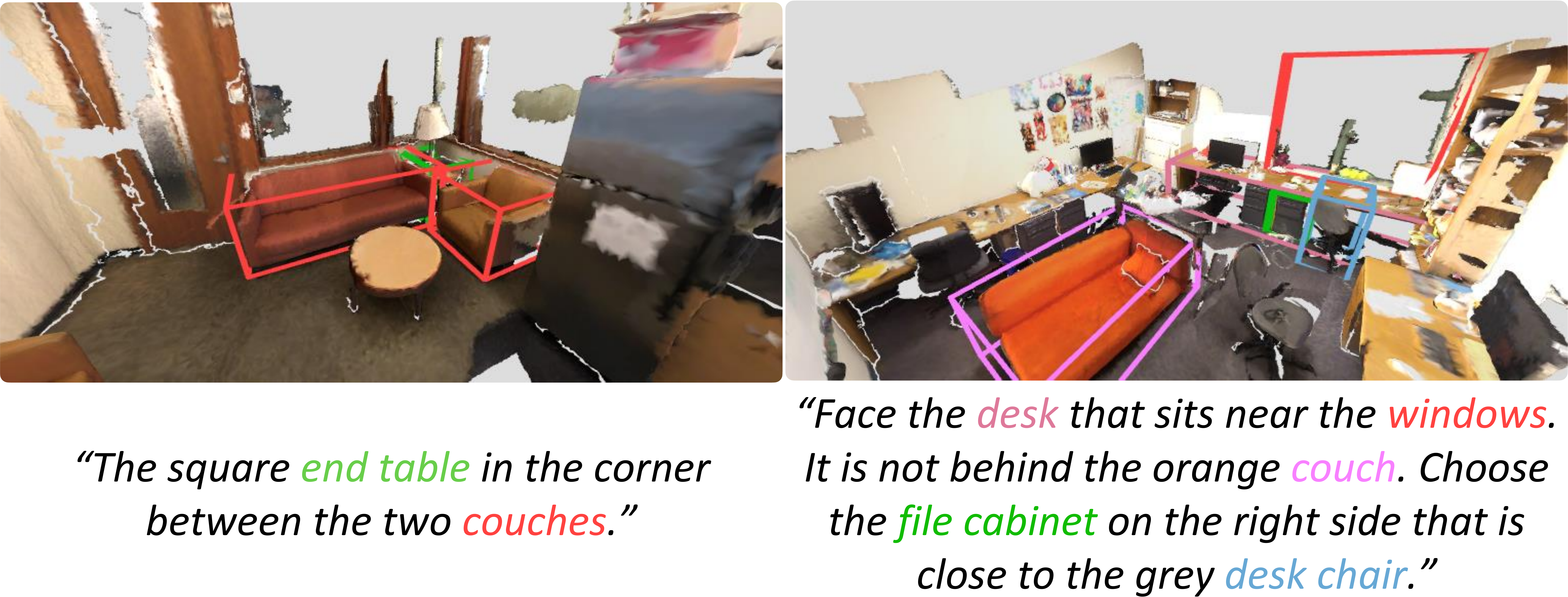} }}%
    \qquad
    \subfloat[\centering Examples of \scanRefer{}]{{\includegraphics[width=0.98\linewidth]{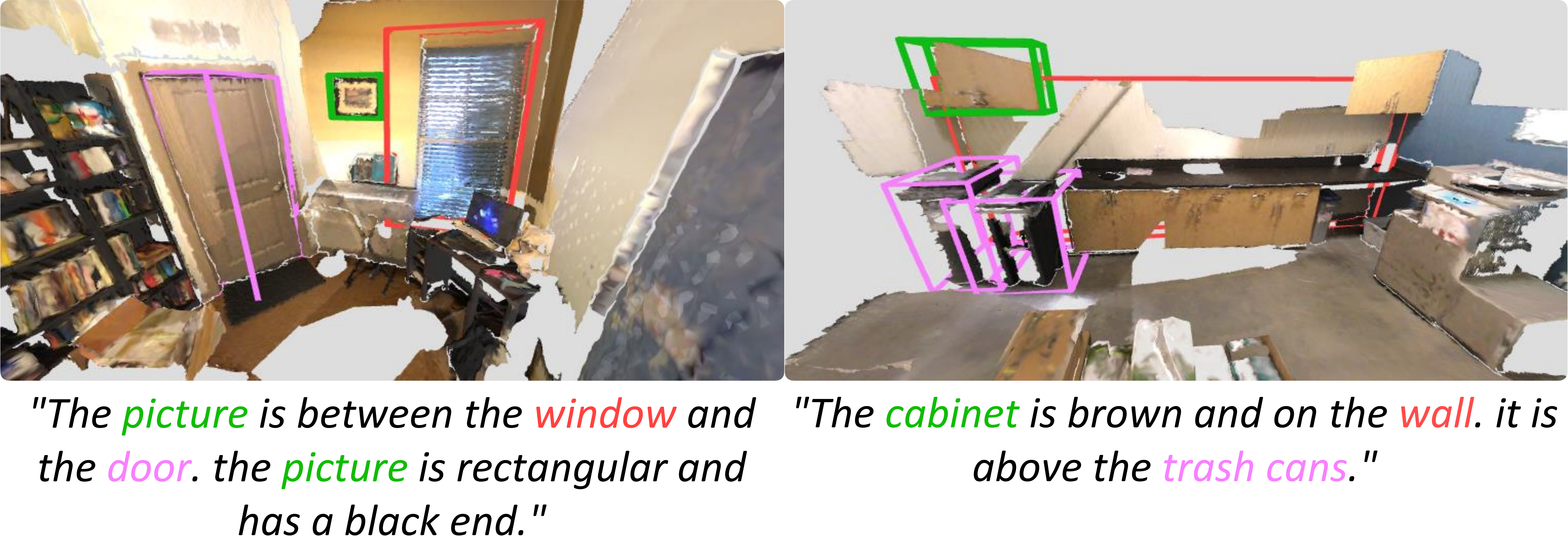} }}
    \caption{\textbf{Typical annotation examples from \datasetName{}}. 
    Our annotations link each noun phrase in a given referential sentence to one or more corresponding objects in a 3D ScanNet scene. The target object and its corresponding noun phrase are shown in green. The anchor objects and their corresponding noun phrases are shown in different colors. The couches on the top left and the trash cans on the bottom right are examples where one noun phrase corresponds to multiple objects in the scene.
    }
    \label{fig:dataset_teaser}
\end{figure}

% As the amount of available data from both the linguistic and 3D domains has increased drastically in recent years, so too has an interest in sophisticated techniques to combine, understand, and exploit this data to solve outstanding problems involving both domains.
Connecting natural language to real-world 3D scenes enables us to tackle fundamental problems such as language-assisted object localization and fine-grained object identification~\cite{zhenyu2019scanrefer,achlioptas2020referit_3d}, object captioning~\cite{chen2021scan2cap}, scene-based Q/A~\cite{Azuma2021ScanQA3Q}, and language-based semantic segmentation~\cite{Rozenberszki2022LanguageGroundedI3}.

\begin{table*}[!htb]
    \centering
\begin{myresizeenv}
\begin{tabular}{lcccccc}
\toprule
  &  \makecell[c]{\#Utterances}  & \makecell[c]{\#Annotated \\ Objects}  & \makecell[c]{Anchor Instance \\ Annotations} & \makecell[c]{Phrase-to-Object \\ Correspondences} & \makecell[c]{\#Scan \\ Entities} &
  \makecell[c]{Avg. \# of Objects \\ per Scan Entity}  \\
\hline
\nr{}~\cite{achlioptas2020referit_3d} &  38K & 38K & \xmark{} & \xmark{} &  - &  - \\
\scanRefer{}~\cite{zhenyu2019scanrefer}  & 46K & 46K  & \xmark{} & \xmark{}  & - &  -\\
\hline
\multicolumn{7}{c}{{\datasetName{}}} \\
\hline
\nr{}-\datasetSuffix{} & 38K & \textbf{126K} \textcolor[RGB]{102,178,84}{(\textbf{+88K})}  & \cmark{}& \cmark{} & \textbf{\numberOfEntities{}} & \textbf{\AverageLengthOfObjectsPerEntity{}}\\
\scanRefer{}-\datasetSuffix{}  & 46K &  \textbf{243K} \textcolor[RGB]{102,178,84}{(\textbf{+197K})}   & \textbf{\cmark{}}& \cmark{} & \textbf{\numberOfEntitiesScanRefer{}} & \textbf{\AverageLengthOfObjectsPerEntityScanRefer} \\
\bottomrule
\end{tabular}
 \end{myresizeenv}
    \caption{\textbf{Comparison between the \nr{} and \scanRefer{} datasets and their corresponding extensions in \datasetName{}.} Our proposed dataset contains more annotated objects and provides annotations for the anchor objects mentioned in the referential utterances. Specifically, \datasetName{} provides explicit phrase-to-object correspondences for \textit{all} mentioned objects. \scanRefer{} has more verbose utterances compared to the more parsimonious \nr{}. This distinction is also reflected in the resulting statistics from \datasetName{} (last two columns).}
    \label{tab:dataset_stats}
\end{table*}

To improve upon these types of problems, we extend two recent datasets \scanRefer{} and \nr{} with a new type of annotation.
These two datasets collected referential sentences for real-world 3D scenes. A referential sentence describes a single (``target") object in a 3D scene. The grounding annotations in these two datasets consist of labeling the target object in the scene and associating it with the referential sentence.
Such a referential sentence needs to discriminate the target object from the remaining objects in the 3D scene.
This can be done by emphasizing properties of the target object such as color, material, or geometry (e.g., \textit{the tall chair}).
However, we can observe that most referential sentences contain information about other objects and object relationships in order to describe the target object (e.g., \textit{the tall chair} $\rightarrow$ \textit{the tall chair between the table and the fireplace}). We call these other objects (``anchor objects").

In our work, we set out to utilize anchor objects in two ways. First, we propose a new dataset \datasetName{}. We curate grounding annotations for \textit{all} 3D objects in each referential sentence for both \nr{} and \scanRefer{}. Previously, grounding annotations were only available by linking a single target object to a complete referential sentence. In contrast, we provide grounding annotations by linking target and anchor objects to noun phrases within the referential sentence. We call this new type of data a \textit{scan entity}. A scan entity consists of phrases (e.g., tables, trash cans) along with the 3D objects of the scene that correspond to them (see \Cref{fig:dataset_teaser}).
Second, we show how this new type of data can benefit language-based 3D scene understanding in two tasks: discriminative language comprehension (`neural listening') and generative language production (`neural speaking'). 
It is important to note that it is not possible to directly utilize our new annotations in existing architectures. We, therefore, propose several architecture modifications and training losses to recent frameworks so we can make use of anchor objects. These modifications will make use of the additional information only during training to facilitate the incorporation of auxiliary losses, but no additional data is used during inference time. The goal of our modifications is to predict the anchor objects in addition to the target object. This idea is based on our hypothesis that 3D visio-linguistic architectures \textit{can and should} model pairwise or higher-order object-to-object relations in order to become more robust learners. 
% We demonstrate this intuitive hypothesis by a variety of experimental results concerning two \textit{cornerstone} tasks for language-based 3D scene understanding: experiments addressing object-centric discriminative language comprehension (a.k.a. `neural listening'), and experiments concerning object-centric generative language production (a.k.a. `neural speaking'). Specifically, we demonstrate that the incorporation of appropriate loss functions that tap into \datasetName{} to disentangle and localize the various objects mentioned in a referential utterance are: 
Our modifications are
i) \textit{effective}, as they result in significantly improved accuracy for both tasks in well-established benchmarks; ii) \textit{robust}, as they have a positive performance effect across many distinct architectures, and iii) their learning effect is intuitive and \textit{interpetable} -- we show that the primary cause of the quantitative gains we attain is learning more and/or better object-to-object relations expressed in the referential language. To summarize, our main contributions are the following:

\begin{itemize}
    \item We introduce a large-scale dataset extending both \nr{} and \scanRefer{} by grounding all objects in a referential sentence. Our \textit{\textbf{\datasetName}} dataset (\underline{Scan} \underline{Ent}itie\underline{s} in 3D) includes 369,039 language-to-object correspondences, more than three times the number from the original works. 
 
    \item We propose novel training losses and architecture modifications to exploit the new annotations. We improve the performance of several 3D neural listening architectures, including improving the SoTA in \nr{} and \scanRefer{} by \textbf{\mainBoost{}}, and \textbf{5.0\%} respectively. We improve neural speaking architectures, as measured with
    standard captioning metrics (e.g., BLEU, METEOR, ROUGE, and CIDEr). For instance, we improve
    the SoTA for neural speaking with \nr{}, per CIDEr, by \textbf{+13.2}. Importantly, we note that  we do \textit{not} train our networks with more referential sentences or use \datasetName's annotations during inference. Instead, we rely on additional grounding information during training only.
        
% \item We present quantitative and qualitative results indicating that by training with \datasetName, different neural speaking, or listening, architectures not only attain improved performance but do so in an intuitive and \textit{interpretable} manner, i.e., by better learning high-order (primarily, spatial) object-to-object relations.
\end{itemize}

We acknowledge two strong concurrent works that share a similar idea~\cite{yuan2022toward,sharma2022denserefer3d}. 
As an advantage of our realization, we: 1) have professional instead of crowd-sourced annotations, 2) are the only ones to tackle neural speaking, 3) tackle both ReferIt3D and ScanRefer setups in neural listening, 4) have the best overall results across widely adopted evaluation metrics, 5) propose and explore zero-shot transfer learning for neural listeners operating in novel 3D scenes~\cite{wald}.
\section{Related Work and Background}

\paragraph{Modern visio-linguistic tasks for objects in 3D scenes.}

Increasingly more tasks involving a joint understanding of computer vision and language processing are been studied thanks to the introduction of modern 3D-oriented datasets~\cite{shapenet,ScanNet,silberman2012indoor,xiao2013sun3d,hua2016scenenn,armeni20163d,reizenstein2021common,Li20223DCC,fu20213d} equipped with linguistic annotations~\cite{chen2018text2shape,achlioptas2019shapeglot,hong2021vlgrammar,SNARE,achlioptas_23_shapetalk}. These include captioning of 3D objects in synthetically generated contexts~\cite{achlioptas2019shapeglot,han2020shapecaptioner} and captioning of objects embedded in real-world scenes~\cite{chen2021scan2cap, Yuan_2022_CVPR}, 3D object identification in scenes~\cite{zhenyu2019scanrefer,achlioptas2020referit_3d, Yuan_2022_CVPR,wang2023text}, language-based semantic segmentation~\cite{hou2021exploring,Rozenberszki2022LanguageGroundedI3,PartGlot}, and 3D question answering~\cite{kolve2017ai2,gordon2018iqa,wijmans2019embodied,yu2019multi,Azuma2021ScanQA3Q,ma2023sqa3d}. Existing visio-linguistic datasets involving objects in real-world 3D scenes~\cite{zhenyu2019scanrefer,achlioptas2020referit_3d} provide limited annotations focusing only on target objects, bypassing all other mentioned context-relevant object instances. Despite that, such limited annotations naturally impede the development of more sophisticated 3D neural listeners, a flourishing line of works is being currently developed, concentrating on neural listening~\cite{achlioptas2020referit_3d,SAT_3D,zhao2021_3DVG_Transformer,LanguageRefer,InstanceRefer,TransRefer3D,huang2022multi,Bakr2022LookAA,chen2022unit3d,wu2022eda}, and neural speaking~\cite{Yuan_2022_CVPR,cai20223djcg,chen2022unit3d,zhong2022contextual}.

\paragraph{3D-based visio-linguistic grounding.}
Visio-linguistic grounding aims at associating information expressed in language, e.g., noun-entities, to the underlying objects present in visual stimuli~\cite{plummer2015flickr30k}. Such grounding for 2D images has been extensively studied~\cite{kazemzadeh2014referitgame,plummer2015flickr30k,mao2016generation,yu2016modeling,yu2018mattnet,yang2019fast,yang2020improving}. On the contrary, 3D visual grounding is still in its infancy~\cite{angel_phd_thesis,achlioptas_phd_thesis,achlioptas_23_shapetalk}. Recently, \scanRefer{}~\cite{zhenyu2019scanrefer} and Referit3D~\cite{achlioptas2020referit_3d} proposed datasets for language-driven neural-based comprehension in 3D, built on top of assets of ScanNet~\cite{ScanNet}. Following these establishments, several approaches explored novel designs and new formulations~\cite{SAT_3D, LanguageRefer,Feng2021FreeformDG,cai20223djcg,Wang2022SpatialityguidedTF, Abdelreheem20223DRefTransformerFO,LADIS} for creating improved neural listeners that \textit{implicitly} attempt to model the grounding (visual) context of each reference~\cite{SAT_3D,zhao2021_3DVG_Transformer,LanguageRefer,InstanceRefer,TransRefer3D,huang2022multi}. By using the explicit annotations provided in \datasetName{}, we take a step in reducing the gap between the richer 2D-based and less mature 3D-based learning-based comprehension paradigms. As we show, by developing appropriate adaptations that take into account \datasetName{}, we can improve neural listeners and neural speakers across many architectural designs, including improving two state-of-the-art methods, SAT~\cite{SAT_3D} and MVT~\cite{huang2022multi}.

\section{\datasetName{} Dataset}
\label{sec:sec_dataset}

\begin{figure}
    \centering
    \includegraphics[width=1.0\linewidth]{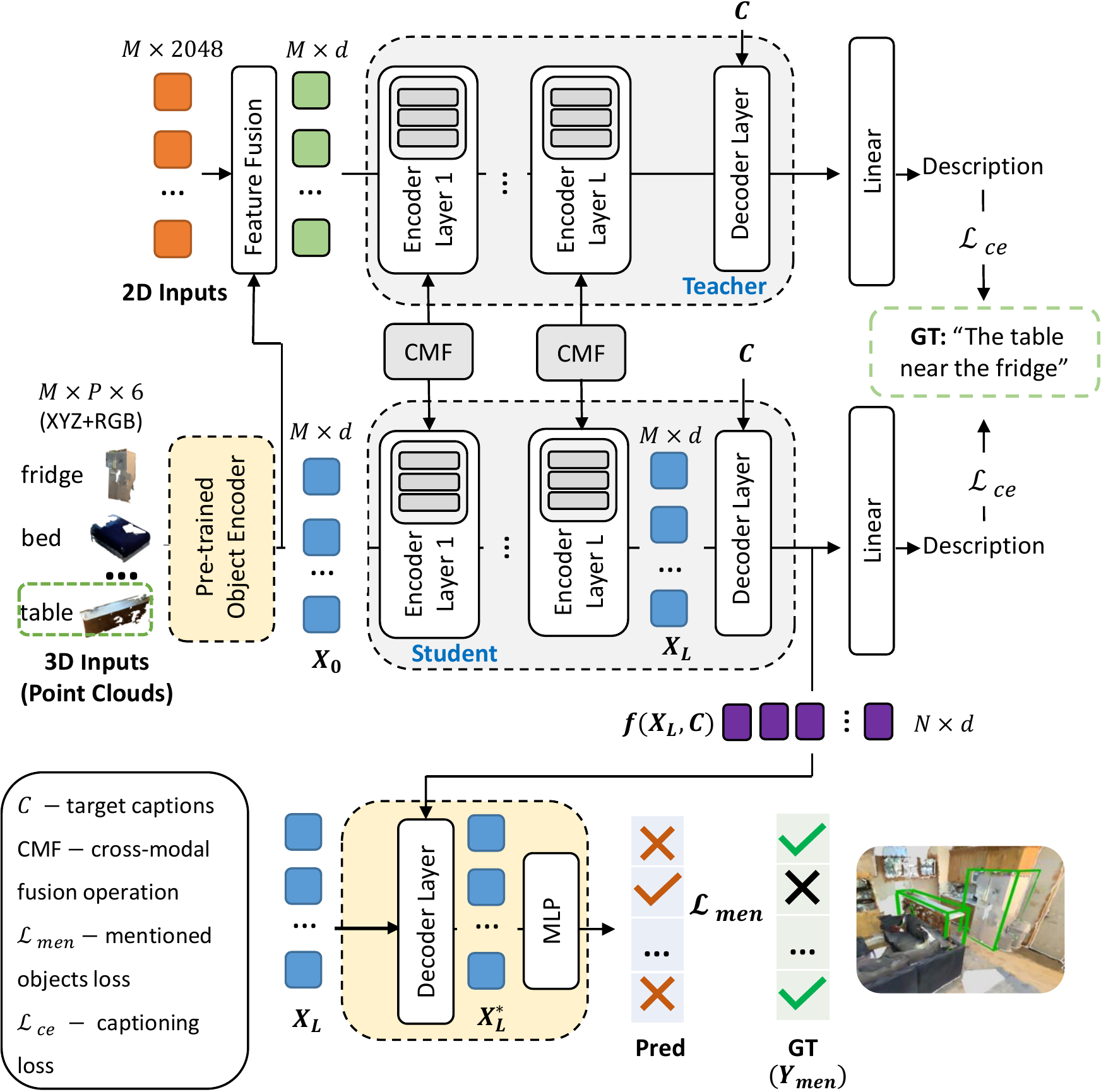}
    \caption{\textbf{Proposed M2Cap-\datasetSuffix{} model adapting $\mathcal{X}-$Trans2Cap model to operate with our proposed losses.} The model is given a set of 3D objects in a 3D scene and outputs a caption for the target object (e.g., table in green box). The $\mathcal{X}-$Trans2Cap model exploits cross-modal knowledge transfer (3D inputs together with their counterpart 2D images) and adopts a student-teacher paradigm~\cite{Chen2017LearningEO,Yuan_2022_CVPR}. Boxes in yellow show our modifications. Here, we use a transfer learning approach by fine-tuning a pre-trained object encoder trained on the listening task to promote discriminative object feature representations. Our modular loss guides the network to predict all object instances mentioned in the ground truth caption.}
    \label{fig:method_m2cap}
\end{figure}

\subsection{Curating human annotations}
Curating all correspondences between each noun phrase in a referential sentence and their underlying objects within a 3D scene is generally an error-prone task.
First, it requires the annotators to be familiar with (albeit simple) linguistic and syntactic rules in the given language to parse the sentence.
Second, they must be able to carefully navigate inside a complicated (and, possibly, poorly reconstructed) scene, which typically contains multiple objects of the same fine-grained object class (e.g., multiple kitchen cabinets, as in the right-most example in \Cref{fig:dataset_teaser}), so as to select \textit{all and only} the correct referenced objects.
In order to ensure the curation of high-quality correspondences with a low error rate and high coverage, we took several critical steps.
First, we developed a custom web-based UI for 3D scene navigation, which was interactive, lightweight (i.e., fast), user-friendly, and which allowed for maintaining an active dialogue with the annotators.
Second, we coordinated with a team of \textit{professional} data labelers to ensure the collection of sufficiently accurate labels for \datasetName{}.

While a common approach to large-scale data collection today is to use crowd-sourcing techniques with platforms such as Amazon Mechanical Turk (AMT)~\cite{10.1007/978-3-642-35142-6_14}, we note that we conducted an AMT-based \textit{pilot} study to determine whether such an approach is sufficient, given the aforementioned complexity and specificity of this task.
We found that the error rate within the collected annotations was significantly higher than that in the annotations provided by the professional labelers (error rates of 16\% vs. $<$ 5\%, respectively).
Rather than attempting to evaluate our approach using data with such a high percentage of erroneous labels, we ultimately decided to employ professional annotators, which significantly improved the attained quality of \datasetName{}.

Finally, we split the curation process into two phases; the annotation phase and the verification phase.
The verification phase also involved \textit{correcting} the mistakes found so as to provide high-quality annotations.
In \Cref{fig:dataset_teaser}, we show examples from the \datasetName{} dataset for \nr{} and \scanRefer{}, which demonstrate that our annotations cover different classes of anchor objects and that our annotations provide rich contexts for these utterances.

\subsection {Key Characteristics of \datasetName{}}
In this section, we briefly present key characteristics of the \datasetName{} dataset. 
% A scan entity associated with a given utterance is a pair of words or short phrases (e.g., \textit{tables}, \textit{trash can})  along with the 3D objects of the scene that correspond to them (see \Cref{fig:dataset_teaser}).
In \Cref{tab:dataset_stats}, we present the number of collected annotations for \numberOfAnnotatedUtterances{} examples from the \nr{} dataset and \scanReferTotalUtterances{} examples from \scanRefer{}. We observe that in general \scanRefer{} annotations provide more entities per single utterance compared to \nr{} (182,300 vs. 96,032, respectively), as \scanRefer{} utterances are typically longer and more verbose than \nr{} utterances (on average, there are 20.3 words per utterance in \scanRefer{}, vs. 11.4 in \nr{}). 

We also calculate how frequently an object is used as an anchor object when it is the \textit{only} 3D instance of its class inside a scene (e.g., the \textit{window} in the lower left example in \Cref{fig:dataset_teaser} ). We find that 24.3\% of all anchor objects are `unique' in \nr{}. However, significantly more anchor objects are unique in \scanRefer{} (39.1\%). Such anchors typically represent \textit{salient} objects~\cite{achlioptas2020referit_3d}, and can be particularly useful for locating the target, esp. when many other objects are being described in context (explaining the differential between the two datasets).

Last, by using our collected annotations, we can extract \textit{object-to-object} spatial relationships of scan entities (with $\sim$91\% verified sampled accuracy), using existing spatial relation classifiers~\cite{nichols-botros-2015-sprl}. Crucially, to attain this accuracy level, we explicitly apply such a classifier on \textit{ground-truth} referred entities found in \datasetName{}. Out of the 13 spatial relationship \textit{types} found, the most frequently used relation in \nr{} and \scanRefer{} is the ``closest" and the ``on top of", respectively. For a more detailed analysis of these findings, we encourage the reader to consult the Supp.

\section{Method}
\begin{figure}
    \centering
    \includegraphics[width=\linewidth]{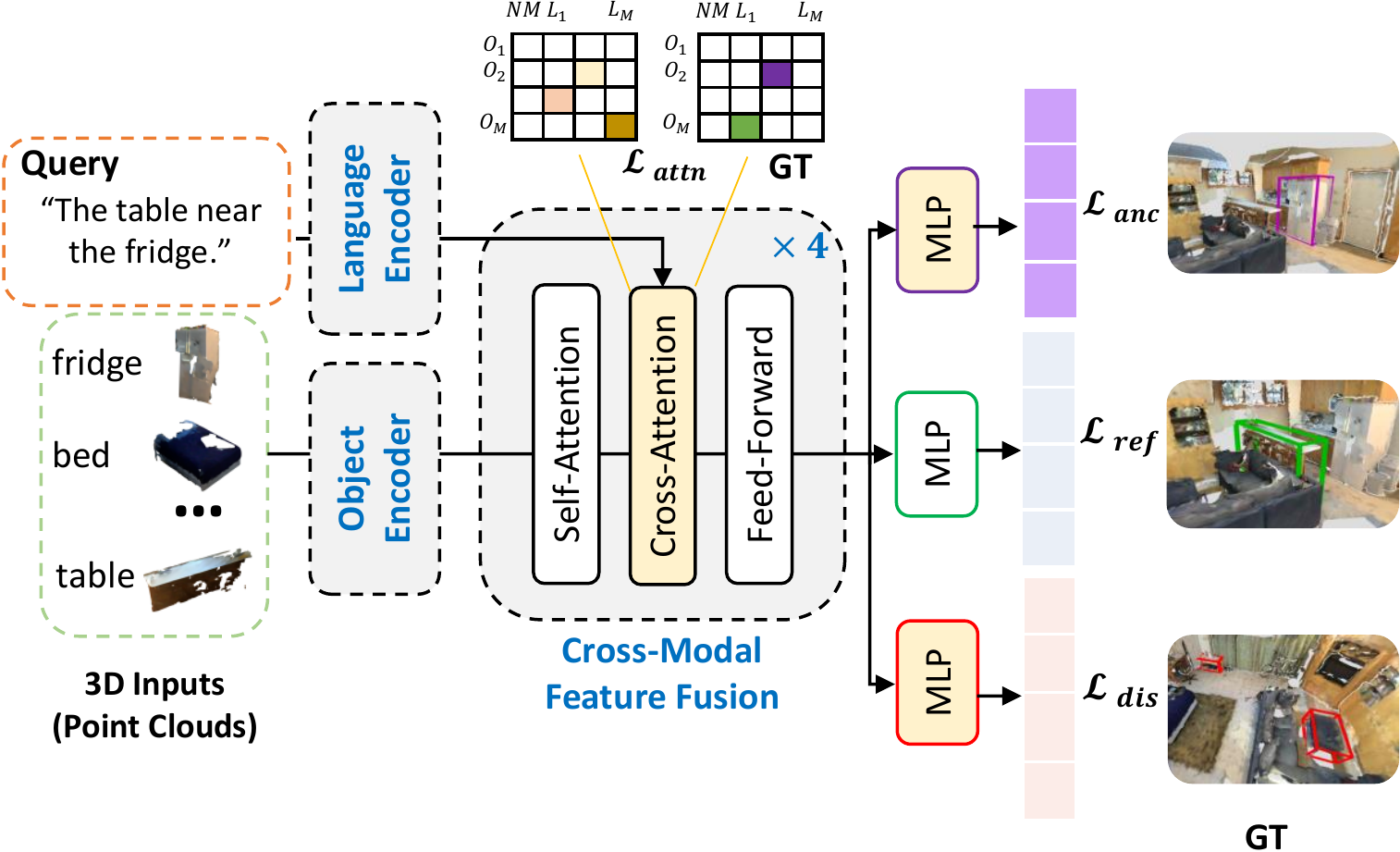}
    \caption{\textbf{Demonstration of our proposed listening losses adjusted for the MVT model.}
    The losses are applied independently of each other on top of object-centric and context-aware features. Crucially, the extended MVT-\datasetSuffix{} model can predict all anchor objects (shown in purple), same-class distractor objects (red), and the target (green). The default model only predicts the target.}
    \label{fig:mvt}
\end{figure}
\label{sec:sec_method}
In this section, we propose modifications to several existing state-of-the-art architectures to utilize the additional annotations provided by \datasetName{} during training. The main idea of the modifications is to use the prediction of anchor objects as an auxiliary task during training. While the exact implementation of this idea depends on the specific architecture, it seems intuitive that an additional understanding of anchor objects will lead to better models. 
We explore two tasks: neural listening and neural speaking and multiple architectures per task. Our main goal is to demonstrate the inherent value of the curated annotations. All proposed modifications are simple to implement and lead to substantial improvements. We, therefore, conjecture that similar modifications are (or will be) possible to existing (and future) architectures making use of \datasetName{}. We also encourage the reader to consult the supplementary material for more details regarding our modifications and their effect.

For neural listeners, we propose three new loss functions. We try these losses on two recent listening architectures, SAT~\cite{SAT_3D} and MVT~\cite{huang2022multi}. In addition, we also propose modifications to 3DJCG~\cite{cai20223djcg} described in the supplementary. For neural speakers, we propose corresponding modifications and appropriate losses for the Show, Attend, and Tell model~\cite{xu2015show} and  $\mathcal{X}-$Trans2Cap model~\cite{Yuan_2022_CVPR}.

% \begin{figure*}[ht!]
%     \centering
%     \includegraphics[width=\linewidth]{Figures/listen_qual-cropped.pdf}\qquad
%     % \includegraphics[width=\linewidth]{Figures/qualitativeResultsNr3d2.jpg}\qquad
%     \caption{\textbf{Qualitative results for our proposed model (MVT-\datasetSuffix{}) compared to the MVT model.} The rows from top to bottom show the ground truth (green box), the target object predicted by MVT (red box), the predicted target object predicted by MVT-\datasetSuffix{} (green box) along with the predicted anchor objects (purple boxes), and the input utterance. The above examples show that the model can accurately predict the target object and simultaneously also predict the underlying anchor objects mentioned in the input utterance.}
%     \label{fig:qualitativeResultsListener}
% \end{figure*}

\subsection {3D Grounded Language Comprehension}
The goal of a neural listener is to identify the target object in a 3D scene described in a referential sentence. Following~\cite{achlioptas2020referit_3d}, the input to our neural listener is a set of $M$ 3D object proposals present in a 3D scene, where each proposal is represented as a 3D point cloud, and an input utterance describing the target object, represented as a sequence of $N$ tokens.
Most recent neural listeners are transformer-based models~\cite{ SAT_3D,zhao2021_3DVG_Transformer,huang2022multi}, each of which applies bi-modal attention between the features of the 3D objects and the features of the words of the input utterance. Assuming this generic setup, we now detail our three proposed loss functions.

\subsubsection{Anchor Prediction Loss}
\label{anc_loss}
The anchor prediction loss $\mathcal{L}_\mathrm{anc}$ guides the neural listener to predict the anchor objects (non-target objects that are mentioned in an input utterance).
In order to identify the target object, one must typically also identify the mentioned anchor objects. The anchor prediction loss can be applied to any output token of an attention or self-attention layer.
We obtain a suitable set of tokens (feature vectors) for the $M$ input 3D object proposals denoted as $F_{O}=\{f_0, f_i,\dots, f_M\}$ as follows. For the MVT model~\cite{huang2022multi}, $F_{O}$ is obtained from a sequence of transformer decoder layers followed by aggregation over multiple views as shown in \Cref{fig:mvt}.
For the SAT model~\cite{SAT_3D}, $F_{O}$ is obtained from a sequence of multi-modal attention layers. We derive $X_\mathrm{anc}=\phi(F_{O})$ with an auxiliary classification head using an MLP to encode $\phi(.)$. The MLP consists of two fully connected layers, where $X_\mathrm{anc}$ represents a vector capturing the listener's confidence of each object being an anchor object (of shape $M\times1$ expressing the logits). We apply a binary cross entropy loss as in \Cref{eq:ancLoss}, where $Y_\mathrm{anc}$ is the ground truth vector of shape $M\times1$. 

\begin{equation}
    \mathcal{L}_\mathrm{anc} = BCE(X_\mathrm{anc}, Y_\mathrm{anc})
    \label{eq:ancLoss}
\end{equation}

\subsubsection{Cross-Attention Map Loss}
The Cross-Attention Map loss encourages the network to attain high relevance values between the objects and the words corresponding to the same underlying scan entity.
This loss operates on cross-attention maps $A$ (before applying the softmax operation) between the features of the input scene 3D objects and the word tokens of the input utterance, where $A$ is of shape $M \times N$.
The target matrix $Y_\mathrm{attn}$ is a binary matrix of shape $M \times N$, where a cell ($y_{i, j}$) has a value of 1 if the $i$th object and the $j$th word correspond to one another.
For each row $R_i$ of shape $1\times N$ and the corresponding row $Y_\mathrm{attn}^{i}$ in the target matrix, the cross-attention map loss ($\mathcal{L}_\mathrm{attn}$) is measured as:
\begin{equation}
\mathcal{L}_\mathrm{attn} = \frac{1}{M} \sum_{i=1}^{M} BCE(R_i, Y_\mathrm{attn}^{i})
\label{eq:attnLoss}
\end{equation}

\begin{table*}[!htb]
\small
    % \begin{myresizeenv}
        \centering
        \begin{tabular}{lccccc}
            \multicolumn{1}{c}{Arch.}  & \multicolumn{1}{c}{Overall} & \multicolumn{1}{c}{Easy} & \multicolumn{1}{c}{Hard} & \multicolumn{1}{c}{View-dep.} & \multicolumn{1}{c}{View-indep.} \\ 
            \toprule
            % \multicolumn{6}{c}{{}} \\ 
            ReferIt3DNet~\cite{achlioptas2020referit_3d}                          & 35.6\%$\pm$0.7\%            & 43.6\%$\pm$0.8\%         & 27.9\%$\pm$0.7\%         & 32.5\%$\pm$0.7\%              & 37.1\%$\pm$0.8\%                \\
            InstanceRefer~\cite{InstanceRefer}                       & 38.8\%$\pm$0.4\%            & 46.0\%$\pm$0.5\%         & 31.8\%$\pm$0.4\%         & 34.5\%$\pm$0.6\%              & 41.9\%$\pm$0.4\%                \\
            3DRefTransformer~\cite{Abdelreheem20223DRefTransformerFO}  & 39.0\%$\pm$0.2\% &  46.4\%$\pm$0.4\% & 32.0\%$\pm$0.3\% & 34.7\%$\pm$0.3\% & 41.2\%$\pm$0.4\% \\
            3DVG-Transformer~\cite{zhao2021_3DVG_Transformer}  & 40.8\%$\pm$0.2\%          & 48.5\%$\pm$0.2\%       & 34.8\%$\pm$0.4\%       & 34.8\%$\pm$0.7\%            & 43.7\%$\pm$0.5\%              \\
            FFL-3DOG~\cite{Feng2021FreeformDG}           & 41.7\%                      & 48.2\%                   & 35.0\%                   & 37.1\%                        & 44.7\%                          \\
            TransRefer3D~\cite{TransRefer3D}     & 42.1\%$\pm$0.2\%            & 48.5\%$\pm$0.2\%         & 36.0\%$\pm$0.4\%         & 36.5\%$\pm$0.6\%              & 44.9\%$\pm$0.3\%                \\
            LanguageRefer~\cite{LanguageRefer}      & 43.9\%                      & 51.0\%                   & 36.6\%                   & 41.7\%                        & 45.0\%                          \\
            
            SAT~\cite{SAT_3D}             & 49.2\%$\pm$0.3\%            & 56.3\%$\pm$0.5\%         & 42.4\%$\pm$0.4\%         & 46.9\%$\pm$0.3\%              & 50.4\%$\pm$0.3\%                \\
            3D-SPS~\cite{luo20223d} & 51.5\%$\pm$0.2\%  & 58.1\%$\pm$0.3\%         & 45.1\%$\pm$0.4\%         & 48.0\%$\pm$0.2\%              & 53.2\%$\pm$0.3\%                \\
            PhraseRefer~\cite{yuan2022toward} & 54.4\% &  62.1\% & 47.0\% & 51.2\% & 56.0\% \\ 
            MVT~\cite{huang2022multi}                                                           & 55.1\%$\pm$0.3\%            & 61.3\%$\pm$0.4\%         & 49.1\%$\pm$0.4\%         & 54.3\%$\pm$0.5\%              & 55.4\%$\pm$0.3\%              \\
            SAT-\datasetSuffix{} (ours)  &
            \makecell[c]{{52.5\%$\pm$0.2\%} \\ \footnotesize \textbf{\textcolor[RGB]{102,178,84}{(+3.3\%)}}} & 
            \makecell[c]{{59.8\%$\pm$0.2\%} \\ \footnotesize \textbf{\textcolor[RGB]{102,178,84}{(+3.6\%)}}} &
            \makecell[c]{{45.6\%$\pm$0.3\%} \\ \footnotesize \textbf{\textcolor[RGB]{102,178,84}{(+3.2\%)}} }&
            \makecell[c]{{51.3\%$\pm$0.5\%} \\ \footnotesize \textbf{\textcolor[RGB]{102,178,84}{(+4.4\%)}}} &  
            \makecell[c]{{53.2\%$\pm$0.1\%} \\ \footnotesize \textbf{\textcolor[RGB]{102,178,84}{(+2.8\%)}}} \\
  
            MVT-\datasetSuffix{} (ours) &
            \makecell[c]{\textbf{59.3\%$\pm$0.1\%} \\ \footnotesize \textbf{\textcolor[RGB]{102,178,84}{(+4.2\%)}}} & 
            \makecell[c]{\textbf{65.4\%$\pm$0.3\%} \\ \footnotesize \textbf{\textcolor[RGB]{102,178,84}{(+4.1\%)}}} & 
            \makecell[c]{\textbf{53.5\%$\pm$0.2\%} \\ \footnotesize \textbf{\textcolor[RGB]{102,178,84}{(+4.4\%)}}} &
            \makecell[c]{\textbf{57.3\%$\pm$0.3\%} \\ \footnotesize \textbf{\textcolor[RGB]{102,178,84}{(+3.0\%)}}} &  
            \makecell[c]{\textbf{60.4\%$\pm$0.2\%} \\ \footnotesize \textbf{\textcolor[RGB]{102,178,84}{(+5.0\%)}}} \\
            
            \bottomrule
        \end{tabular}
        \caption{\textbf{Listening performance on \nr{} dataset.} The neural listeners are trained with or without our proposed \nr{} -\datasetSuffix{} dataset and our proposed losses. The numbers in green are the relative improvements over their original counterparts.}
        \label{tab:listening_nr3d}
    % \end{myresizeenv}
\end{table*}

\subsubsection{Same-Class Distractor Prediction Loss}
This loss guides the neural listener to predict the same-class distractor objects ($\mathcal{L}_\mathrm{dis}$). It does not directly leverage \datasetName{} but as we show it offers beneficial synergies with the above losses as it helps to better disentangle the target from distracting objects with the same (fine-grained) object class. Such same-class distractors are objects from the same class as the target co-existing in the scene. As with the anchor objects, we treat the same-class distractor prediction problem as a multi-label classification problem. Thus, we use an approach similar to \Cref{anc_loss}. Specifically, we obtain the logits for predicting the same-class distractor $X_\mathrm{dis} =\psi(F_{O})$ of shape $M\times1$ with an MLP $\psi(.)$. This loss is also binary cross entropy-based, like in \Cref{eq:disLoss}, where $Y_\mathrm{dis}$ is a multi-hot target vector of shape $M\times1$. Note that a same-class distractor object may not be mentioned in the given input utterance. 

\begin{equation}
    \mathcal{L}_\mathrm{dis} = BCE(X_\mathrm{dis}, Y_\mathrm{dis})
    \label{eq:disLoss}
\end{equation}

\subsubsection {Training Objective Function}
The proposed losses can serve as auxiliary add-ons to the original loss term ($ \mathcal{L}_\mathrm{org}$) of existing neural listeners, such as the MVT and SAT models. 
We train these models in an end-to-end fashion as:

\begin{equation}
    \mathcal{L} = \mathcal{L}_\mathrm{org} + \mathcal{L}_\mathrm{aux}, \quad\text{where}\quad  \mathcal{L}_\mathrm{aux} = \alpha \mathcal{L}_\mathrm{anc} + \beta \mathcal{L}_\mathrm{attn} + \gamma \mathcal{L}_\mathrm{dis}
    \label{eq:trainingObjective}
\end{equation}

Where $\alpha$, $\beta$, and $\gamma$ are scalar values controlling the relevant importance of each term. In our experiments, we use $\alpha=\beta=3.0$ and $\gamma=0.5$.

\subsection{Grounded Language Production in 3D}

We describe our modifications to existing architectures for neural speaking. We call our versions of these architectures SATCap-\datasetSuffix{} and  M2Cap-\datasetSuffix{}.

\subsubsection{SATCap-\datasetSuffix{}}
The ``Show, Attend, and Tell" model is an encoder-decoder network originally designed for 2D-based image captioning. To make it amenable to purely 3D inputs, we replace the image encoder with the encoder network found in the MVT model~\cite{huang2022multi}, which is a point cloud PointNet++ encoder together with 3D object self-attention layers. Crucially, to improve the generalization of this speaker, we use a
\textit{pretrained} MVT-based encoder solving the neural-listening task and then fine-tune it for the speaking task. For the decoder network, we use a unidirectional LSTM cell~\cite{lstm}. The encoder part is given the ground-truth objects as input in a similar manner to~\cite{Yuan_2022_CVPR}. The speaker model is trained via teacher-forcing~\cite{teacher_forcing}. Importantly, we also apply our proposed entity prediction loss during the decoding steps. At each decoding step, if the current word to be predicted corresponds to a scan entity, our  loss pushes the object corresponding to the underlying scan entity to be the highest scoring among all objects present in the input scene.

\subsubsection{M2Cap-\datasetSuffix{}}

We employ a similar approach on the $\mathcal{X}-$Trans2Cap model~\cite{Yuan_2022_CVPR}, referred to as M2Cap-\datasetSuffix{} detailed in \Cref{fig:method_m2cap}. We introduce the following two changes to the $\mathcal{X}-$Trans2Cap architecture. First, we use a pre-trained PointNet++ encoder followed by the pre-trained 3D object self-attention layers in the MVT~\cite{huang2022multi} network. Second, we add a new cross-attention layer after the captioning layer found in the student network. 
The layer applies a cross-attention operation between the features of the 3D objects $X_{L}$ of shape $M \times d$ and the features of the predicted tokens $N \times d$, where $d$ is the latent feature dimension, to obtain new enhanced features $X_{L}^{*}$ of shape $M \times d$. Finally, the logit vector is obtained with an MLP $\theta(.)$, representing a confidence value for each object as to whether it is mentioned in the target caption. A binary cross-entropy loss $\mathcal{L}_\mathrm{men} = BCE(\theta{(X_{L}^{*})}, Y_\mathrm{men})$ is employed, in which the target vector $Y_\mathrm{men}$ is a multi-hot vector ($y_\mathrm{men}^{i}$ is $1$ if the $i$th object is mentioned in the target caption). We do not train a speaker and a listener jointly, which is the key contribution of D3Net~\cite{chen2021d3net}. Instead, our focus is on the introduction and utilization of dense annotations.

\begin{table*}[tbh!]
% \begin{myresizeenv}
    \centering
% \begin{adjustbox}{width=0.75\linewidth,center}
\small
\begin{tabular}{lcccccccc}
\multicolumn{1}{c}{Arch.} & 
\multicolumn{4}{c}{\nr{}} & \multicolumn{4}{c}{\scanRefer{}} \\
\toprule
   & C  &  B-4 &  M &  R &  C &  B-4 &  M &  R \\
\hline
Scan2Cap\cite{chen2021scan2cap} &
61.89 & 
32.02 &
28.88 &
64.17 &
64.44 &
36.89 & 
28.42 & 
60.42 \\
$\mathcal{X}-$Trans2Cap \cite{Yuan_2022_CVPR} &
80.02 &
{37.90} &
{30.48} &
{67.64} &
{87.09} &
44.12 &
30.67 &
64.37 \\
\hline
SATCap (ours) & 
76.57&
29.12& 
24.97&
55.62 &
80.98&
37.47&
26.91&
56.98 \\
SATCap-\datasetSuffix{} (ours)
&
{84.37} &
{30.73} &
{25.90} &
{56.57} &
{84.81} &
{38.85} &
{27.18} &
{57.62}  \\
\hline
M2Cap (ours) 
&
86.15 &
37.03 &
30.63 &
67.00 &
85.75 &
44.02 & 
30.74 &
64.80 \\
M2Cap-\datasetSuffix{} (ours) &
\textbf{93.25} &
\textbf{39.33} &
\textbf{31.55} &
\textbf{68.33} &
\textbf{87.20} &
\textbf{44.81} &
\textbf{30.93} &
\textbf{65.24} \\
\bottomrule
\end{tabular}
    \caption{\textbf{Speaking performance on \nr{} and \scanRefer{} datasets.} The results of incorporating \datasetName{} dataset in our proposed approaches for the speaking (captioning) task. A speaking model trained with our rich annotations performs better than one trained without them for both the \nr{} and \scanRefer{} datasets.}
    \label{tab:speaking_baseline}
% \end{myresizeenv}
% \end{adjustbox}
\end{table*}

\section{Experiments}
\label{sec:sec_experiments}
\subsection{Experimental Setup}
\begin{table}[!htb]
    \centering
    
    \begin{myresizeenv}
        \begin{tabular}{lcccccc}
        \multicolumn{1}{c}{} & 
        \multicolumn{2}{c}{Unique} &
        \multicolumn{2}{c}{Multiple} &
        \multicolumn{2}{c}{Overall} \\
        \hline
        & \multicolumn{2}{c}{Acc.} &
        \multicolumn{2}{c}{Acc.} &
        \multicolumn{2}{c}{Acc.} \\
        
        & \makecell[c]{@0.25} &
        \makecell[c]{@0.5} &
        \makecell[c]{@0.25} &
        \makecell[c]{@0.5} &
        \makecell[c]{@0.25} &
        \makecell[c]{@0.5}  \\
        \hline
        3DJCG~\cite{cai20223djcg} & 78.75 &\textbf{ 61.30} & 40.13 & 30.08 & 47.62 & 36.14 \\
        3DJCG-\datasetSuffix{} (ours) & \textbf{79.49} & 60.74 & \textbf{41.51} & \textbf{31.34} & \textbf{48.88} & \textbf{37.04} \\
        \bottomrule
        \end{tabular}
    \end{myresizeenv}
    
    \caption{\textbf{Effect of ScanEnts3D for object detector-based listeners}. This ablation shows the effectiveness of using \datasetName{} on a different listener design (\scanRefer{} setup). The attained performance boost further suggests the usefulness and generality of the \datasetName{}-induced loss functions.}
    \label{tab:ablation_differentListeningDesign}
    % \vspace{-3mm}
\end{table}
\textbf{Datasets and splits.} 
We use the \nr{}~\cite{achlioptas2020referit_3d} and \scanRefer{}~\cite{zhenyu2019scanrefer} datasets with their original annotations as well as our additional annotations provided with the proposed \datasetName{} dataset. We use the official ScanNet~\cite{ScanNet} training and validation splits.

\textbf{Metrics.}
For the neural listening experiments, we report the attained target referential accuracy. For the neural speaking experiments we evaluate the output text generations against the ground-truth annotations, based on the metrics of BLEU-4~\cite{Papineni2002BleuAM}, ROUGE~\cite{Lin2004ROUGEAP}, METEOR~\cite{Banerjee2005METEORAA}, and CIDEr~\cite{Vedantam2015CIDErCI}. 

We show the most important results in the paper and leave additional zero-shot tests for the supplementary.

\subsection{Neural Listening}
We demonstrate the effectiveness of the proposed \datasetName{} by comparing state-of-the-art models trained with and without the additional annotations. 
For all experiments, we note that our dataset only leads to modifications at training time. At inference time, our trained models and their respective baseline models use the same input data.

\textbf{Neural listeners trained with \datasetName{} achieve state-of-the-art performance.}
As shown in \Cref{tab:listening_nr3d} and \Cref{tab:listening_scanrefer}, our {MVT-\datasetSuffix{}} neural listener, which is trained with our proposed dataset (\nr{}-\datasetSuffix{}) and our auxiliary losses, achieves state-of-the-art results, outperforming the current SoTA models.
MVT-\datasetSuffix{} outperforms the original MVT~\cite{huang2022multi} on both the \nr{} (+\mainBoost{}) and the \scanRefer{} (+\scanReferMainBoost{}) datasets, while the {SAT-\datasetSuffix{}} model similarly outperforms the original SAT~\cite{SAT_3D} model on both the \nr{} (+3.3\%) and \scanRefer{} (+2.4\%) datasets. 

\textbf{Further analysis.}
Furthermore, we observe considerable improvements in each context for \nr{}, particularly in the view-independent and hard contexts (5.0\% and 4.4\% as in \Cref{tab:listening_nr3d}, respectively). In addition, we report the $F_1$ score~\cite{DBLP:journals/corr/abs-2010-16061}, which measures the overall accuracy of a test taking into account its precision and recall, of the anchor object classification in the MVT-\datasetSuffix{} model. The $F_1$ score of 0.64 (out of a possible maximum of 1) suggests that the full potential value of our proposed dataset \datasetName{} may still be attained with the development of more sophisticated losses, a promising area for future work.

\begin{figure*}[ht!]
    \centering
    \includegraphics[width=\linewidth]{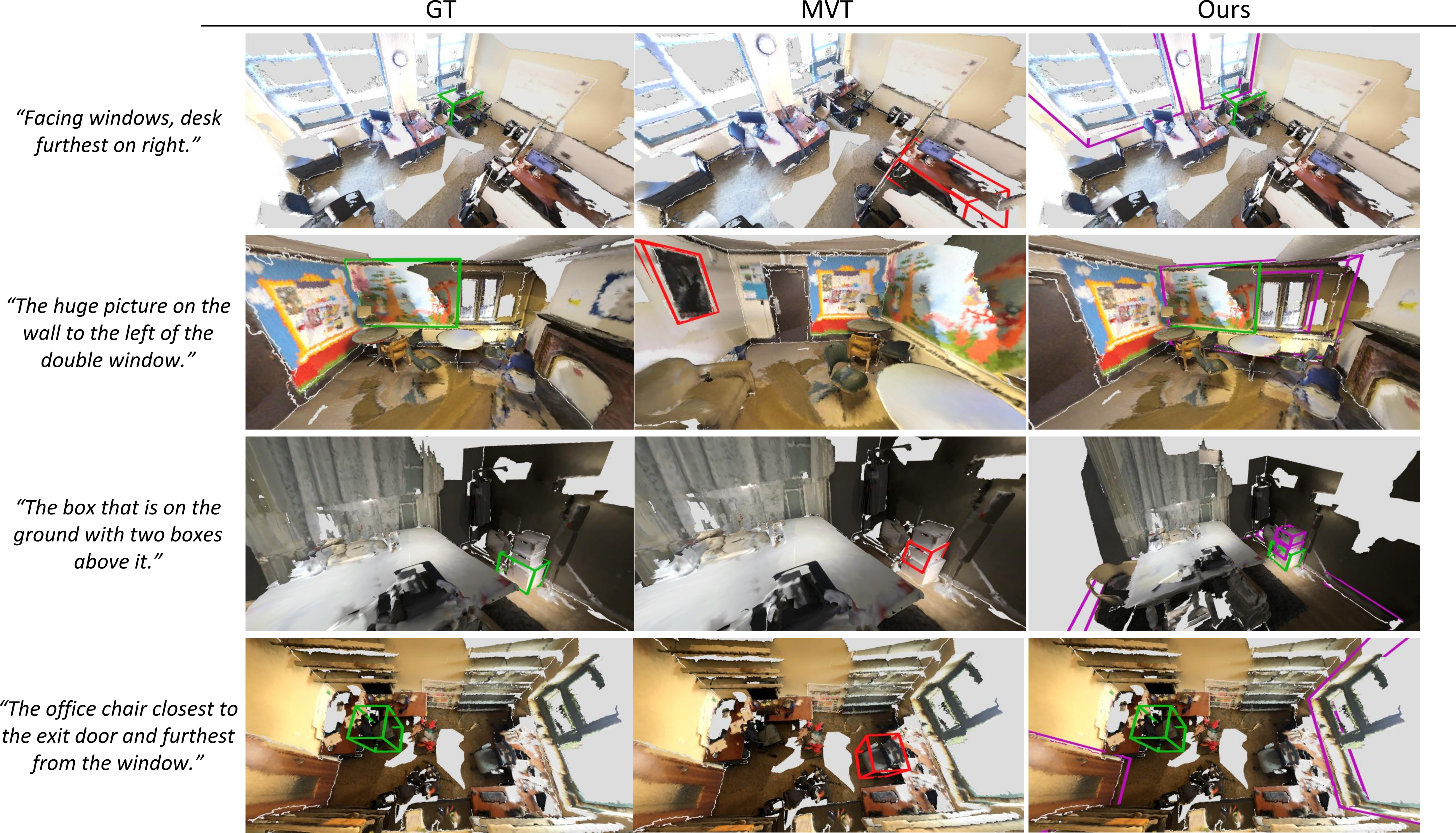}\qquad
    \caption{\textbf{Qualitative results for our proposed model (MVT-\datasetSuffix{}) compared to the MVT model.} The rows from top to bottom show the ground truth (green box), the target object predicted by MVT (red box), the predicted target object predicted by MVT-\datasetSuffix{} (green box) along with the predicted anchor objects (purple boxes), and the input utterance. The above examples show that the model can accurately predict the target object and simultaneously also predict the underlying anchor objects mentioned in the input utterance.}
    \label{fig:qualitativeResultsListener}
\end{figure*}

Finally, in \Cref{fig:qualitativeResultsListener}, we present qualitative examples of how recognizing the anchor objects allows the model to identify the target object correctly. 
Comparing the proposed model MVT-\datasetSuffix{} and the current state-of-the-art method MVT, we demonstrate that guiding our network to understand the anchor entities mentioned in the input utterances promotes the listener to accurately identify the target object. In the third column of this Figure, we demonstrate the predicted target object and the predicted anchor objects by MVT-\datasetSuffix{} in green and purple bounding boxes, respectively.

\begin{table*}
\RawFloats
\centering
\makebox[0pt][c]{\parbox{1\textwidth}{%
    \begin{minipage}[c]{0.67\hsize}
    \centering
    \begin{myresizeenv}
      \begin{tabular}{ccccccccc}
    \\
    \multicolumn{1}{c}{$\mathcal{L}_{attn}$} & \multicolumn{1}{c}{$\mathcal{L}_{anc}$}& \multicolumn{1}{c}{{$\mathcal{L}_{dis}$}} & \multicolumn{1}{c}{{Overall}} & \multicolumn{1}{c}{Easy} & \multicolumn{1}{c}{Hard}& \multicolumn{1}{c}{View-dep.}&\multicolumn{1}{c}{View-indep.} & 
    \\
     \toprule
      & &  &  
        55.1\% & 
        61.3\% & 
        49.1\% & 
        54.3\% & 
        55.4\% &  
        \\ 
     \checkmark & & &  
        56.6\% &  
        63.0\% &  
        50.5\% &  
        55.4\% &  
        57.2\% &
        \\ 
     & & \checkmark &  
        56.9\% &  
        63.5\% &  
        50.6\% &  
        55.3\% &  
        57.8\% &
        \\ 

    \checkmark & & \checkmark &  
        57.4\% &  
        64.3\% &  
        50.8\% &  
        55.6\% &  
        58.3\% &
        \\ 
    & \checkmark & \checkmark &  
        57.9\% &  
        63.7\% &  
        52.3\% &  
        56.0\% &  
        58.9\% &
        \\ 
     & \checkmark & &  
        58.1\% &  
        63.8\% &  
        52.6\% &  
        56.7\% &  
        58.8\% &
        \\ 
    \checkmark & \checkmark & &  
        58.7\%&  
        64.6\%&  
        53.1\%&  
        \textbf{57.5\%} &  
        59.3\% &
        \\ 
    \checkmark & \checkmark & \checkmark &  
        \textbf{59.3\%} &  
        \textbf{65.4\%} &  
        \textbf{53.5\%} &  
        57.3\% &  
        \textbf{60.4\%} &
        \\ 
    \bottomrule
    \end{tabular}
    \end{myresizeenv}
    \caption{\textbf{Ablation study of loss functions.} We ablate different combinations of our proposed auxiliary losses on the MVT neural listener, trained on \nr{} using \datasetName{}.}
    \label{tab:ablationOnLosses}
    \end{minipage}
    \hfil
    \begin{minipage}[c]{0.29\hsize}\centering
    %   \begin{adjustbox}{width=0.65\columnwidth,center}
    \centering
    \begin{myresizeenv}
    \begin{tabular}{lc}
        \multicolumn{1}{c}{Arch.} & \multicolumn{1}{c}{Acc.} \\ 
        \toprule
        \scanRefer{} \cite{zhenyu2019scanrefer}&  44.5\% \\
        ReferIt3DNet \cite{achlioptas2020referit_3d}&  46.9\%$\pm$0.2\% \\
        SAT\cite{SAT_3D}    & 53.8\%$\pm$0.1\%  \\
        MVT\cite{huang2022multi}   & 54.8\%$\pm$0.1\% \\
        \\
        SAT-\datasetSuffix{} (ours)    &
        \makecell[c]{
            56.2\%$\pm$0.2\%
        } \\
        MVT-\datasetSuffix{} (ours) &
        \makecell[c]{
            \textbf{60.8\%$\pm$0.2\%} 
        } \\ 
        \bottomrule
    \end{tabular}
    \end{myresizeenv}
    \caption{\textbf{Listening performance on the \scanRefer{} dataset.} The neural listeners are trained using the ground truth boxes as input with or without using the additional annotations from the \datasetName{} dataset and our proposed losses.}
    \label{tab:listening_scanrefer}
% \end{adjustbox}
    \end{minipage}
}}
% \vspace{-3mm}
\end{table*}

\begin{figure}[!htb]
    \centering
    \includegraphics[width=\linewidth]{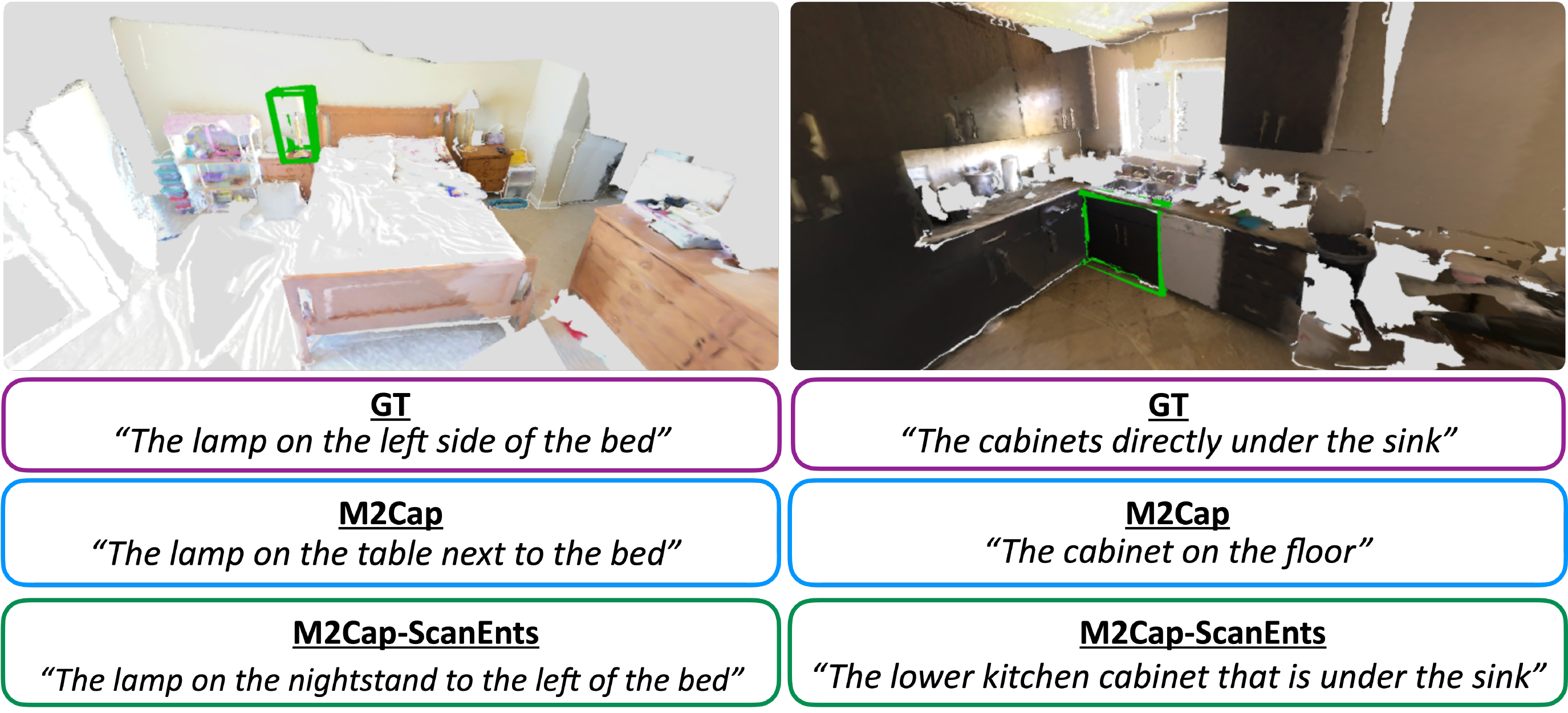}
    \caption{\textbf{ Qualitative comparison of neural speaker variants.} The M2Cap-\datasetSuffix{} generations tend to be more discriminative (e.g., \textit{left of the bed vs. next to the bed}) compared to the default M2Cap variant. In addition, M2Cap-\datasetSuffix{} makes better use of relationships between the target object and anchor objects (\textit{cabinet, sink}).}
    \label{fig:qual_speaker}
\end{figure}

\textbf{Neural listeners trained with \datasetName{} are more context aware.} To show this, first, we conduct additional experiments on both MVT and MVT-\datasetSuffix{} neural listeners  (\Cref{tab:knockoutExperiments}). In these experiments, we change the input to the neural listeners in multiple ways to investigate if the listener becomes better at relying on the context of the 3D scene to  robustly (and more naturally) predict a target object. The changes to the input are the following: (a) an input scene \textit{without} the 3D object proposals of the anchor objects, (b) an input scene with \textit{only} the object proposals of the anchor objects and the same-class distractor objects, and (c) an input utterance where the words that correspond to the anchor objects are replaced with the $<$unk$>$ token denoting an out-of-vocabulary word. We observe that removing the object proposals that correspond to the anchor objects from the input scene results in a massive drop in the listening performance in MVT-\datasetSuffix{}. The drop in the performance in MVT-\datasetSuffix{} is much higher than the drop found in the original MVT model (-15.3\% vs. -7\%). This result suggests that the neural listeners trained with \datasetName{} similar to humans, learn to rely heavily on the anchor objects to identify the target object and are less influenced by the non-anchor/mentioned objects. At the same time, we also observe an improved performance for MVT-\datasetSuffix{} compared to MVT (70.5\% vs. 67.0\%) when providing as input a 3D scene consisting of only the target object, its same-class distractors (to keep the problem highly challenging), and the anchor objects. In other words, on references where humans depend on information about anchors to communicate the target object in a unique manner, we find that visual information about these anchors is both {\em necessary and sufficient} for the performance of our model.

\begin{table}[!htb]% automatically uses minimumwidth
\begin{myresizeenv}
\centering
    \begin{tabular}{lllc}
         \multicolumn{1}{c}{Arch.} &
         \makecell[c]{Anchor Objects \\ Lesioned ($\downarrow$)} &
         \makecell[c]{Anchor Words \\ Lesioned ($\downarrow$)}  &
         \makecell[c]{Anchor Info \\ Present ($\uparrow$)}  \\
         \toprule
         MVT &  48.1\% \textcolor[RGB]{102,178,84}{(-7\%)} &
         45.5\% \textcolor[RGB]{102,178,84}{(-9.6\%)}  &
         67.0\%
         \\
         MVT-\datasetSuffix{} (ours) & \textbf{44.0\%} \textbf{\textcolor[RGB]{102,178,84}{(-15.3\%)}}  & \textbf{44.3\%}  \textbf{\textcolor[RGB]{102,178,84}{(-15.0\%)}} & \textbf{70.5\%}
         \\ 
         \bottomrule
    \end{tabular}
    \caption{\textbf{Evaluating the anchor-object-\textit{awareness} for neural listeners trained w/ and w/o \datasetName{}}. A listener trained with \datasetName{} (MVT-\datasetSuffix{}) learns to depend heavily on the mentioned anchor objects, similar to humans. As seen here, its performance accuracy drops significantly ($\sim$15\%) when the anchors are lesioned from the underlying input, \textit{and} at the same time, its performance gets boosted when only the anchor(s) and the objects of the same class as the target are provided as input.}
    \label{tab:knockoutExperiments}
\end{myresizeenv}
% \vspace{-3mm}
\end{table}
% \vspace{-6mm}

\subsection{Neural Speaking}
With the proposed \datasetName{} dataset, the modified speaker models, SATCap-\datasetSuffix{} and M2Cap-\datasetSuffix{}, improve significantly against their corresponding baseline, as shown in \Cref{tab:speaking_baseline}. 
The encoder networks in SATCap and M2Cap models use the pre-trained encoder weights of an original MVT neural listener trained without \datasetName{}, while the encoder networks in SATCap-\datasetSuffix{} and M2Cap-\datasetSuffix{} use the pre-trained weights of an MVT-\datasetSuffix{} listener. We observe that \datasetName{} helps our speaker models to provide better captions for \nr{} and \scanRefer{} across all metrics (BLEU, CIDEr, METEOR, and ROUGE). The M2Cap-\datasetSuffix{} model improves the SoTA for neural speaking with \nr{}, per CIDEr, by \textbf{+13.2}. In all experiments, we use the ground truth instances as input. Also, we do not provide an extra 2D modality during the inference phase and do not use the additional CIDEr-based loss in the final objective function as in~\cite{Yang_2020_CVPR}. In \Cref{fig:qual_speaker}, we show captions by M2Cap-\datasetSuffix{} on the \nr{} dataset; we compare these captions to those generated by the M2Cap model. We observe that the captions generated by M2Cap-\datasetSuffix{} tend to be more discriminative and make explicit use of valid anchor objects to achieve this desideratum. We refer the reader to the Supp. for ablations on M2Cap-ScanEnts.

\subsection{Ablation Studies}
\textbf{Effectiveness of the proposed losses.} We conduct an ablation study for neural listeners by applying different combinations of our proposed losses. We try each possible combination of our losses ($\mathcal{L}_\mathrm{anc}$, $\mathcal{L}_\mathrm{attn}$, and $\mathcal{L}_\mathrm{dis}$) with the MVT~\cite{huang2022multi} architecture and report their performance on the \nr{} dataset, as shown in \Cref{tab:ablationOnLosses}. When applying $\mathcal{L}_\mathrm{attn}$ alone, we obtain an overall boost of 1.5\% over the baseline MVT model (using none of our proposed losses).
% As mentioned earlier, the boost is considered small for the following reason: forcing the cross-attention maps to be closer to the sparse ground truth matrix prevents the attention layer from attending to other important words (like prepositions, spatial relation words, and object attributes).
% We observe that incorporating the same-class distractor prediction loss helps in improving the referential performance.
We obtain an improvement of 1.8\% upon applying $\mathcal{L}_\mathrm{dis}$ alone. This result is unsurprising, as we find that the same-class distractors are mentioned in 17.2\% of the utterances in the \nr{} and 12.4\% in the \scanRefer{} datasets. Applying the $\mathcal{L}_\mathrm{anc}$ provides the best boost in every experiment where it is applied compared to the other losses. We observe that incorporating the anchor prediction loss is useful for all the \nr{} contexts, especially the hard contexts. The aforementioned result demonstrates how useful the knowledge of the anchor objects mentioned in the input sentence is. The best-performing model applies all three losses, and the performance is better than using $\mathcal{L}_\mathrm{anc}$ and $\mathcal{L}_\mathrm{dis}$ together by 0.6\%.

\textbf{Can \datasetName{} improve 3D object detector-based methods?} As a last experiment, we investigate the extent to which our proposed dataset can improve the performance of different types of neural listeners. In particular, a widely used design proposed by \scanRefer{}~\cite{zhenyu2019scanrefer} requires a listener to first \textit{predict} 3D object proposals and then identify the target object (i.e., 3D object localization). To that end, we adapt the anchor prediction loss to work with the recent 3DJCG network~\cite{cai20223djcg}. In \Cref{tab:ablation_differentListeningDesign}, we see attained improvements in the 3D object localization performance when using our \datasetName{}. Most importantly, we can observe an improvement in the more complex and harder cases (Multiple).

\section{Conclusion}
% Humans describe objects in 3-dimensional environments by understanding and utilizing their relationships with other, co-existing objects.
This work takes substantial initial steps to bring object-to-object interactions, \textit{grounded in language}, to the frontline of relevant learning-based methods.
First, we curate and share a set of rich correspondences covering all referential entities mentioned in \nr{} and \scanRefer{}. Second, we use these annotations to train neural networks with better generalization and understanding of 3D objects w.r.t. their language-based grounding. By adapting existing methods and integrating our proposed loss functions, we attain \textit{SoTA} results in both neural listening and speaking tasks for real-world scenes. We expect the derived insights to open new opportunities to advance related multimodal 3D object-centric tasks. 
% These could include improving the ability of neural agents (e.g., robots) to understand and interact with objects in real-world environments and perform complex language-driven tasks. Describing and manipulating virtual environments (e.g., for captioning or designing virtual worlds) or performing related functions for which proper grounding of 3D entities is crucial for augmenting computers with human-like capabilities.

% \clearpage

%%%%%%%%% REFERENCES
{\small
\bibliographystyle{ieee_fullname}
\bibliography{egbib}

\begin{thebibliography}{10}\itemsep=-1pt

\bibitem{Abdelreheem20223DRefTransformerFO}
Ahmed Abdelreheem, Ujjwal Upadhyay, Ivan Skorokhodov, Rawan~Al Yahya, Jun Chen,
  and Mohamed Elhoseiny.
\newblock {3DRefTransformer}: Fine-grained object identification in real-world
  scenes using natural language.
\newblock {\em WACV}, 2022.

\bibitem{achlioptas_phd_thesis}
Panos Achlioptas.
\newblock {\em Learning to generate and differentiate 3D objects using geometry
  \& language}.
\newblock PhD thesis, Stanford University, 2021.

\bibitem{achlioptas2020referit_3d}
Panos Achlioptas, Ahmed Abdelreheem, Fei Xia, Mohamed Elhoseiny, and
  Leonidas~J. Guibas.
\newblock {ReferIt3D}: Neural listeners for fine-grained 3d object
  identification in real-world scenes.
\newblock In {\em ECCV}, 2020.

\bibitem{achlioptas2019shapeglot}
Panos Achlioptas, Judy Fan, Robert~XD Hawkins, Noah~D Goodman, and Leonidas~J.
  Guibas.
\newblock {ShapeGlot}: Learning language for shape differentiation.
\newblock In {\em ICCV}, 2019.

\bibitem{achlioptas_23_shapetalk}
Panos Achlioptas, Ian Huang, Minhyuk Sung, Sergey Tulyakov, and Leonidas
  Guibas.
\newblock {ShapeTalk}: A language dataset and framework for 3d shape edits and
  deformations.
\newblock {\em IEEE Conference on Computer Vision and Pattern Recognition
  (CVPR)}, 2023.

\bibitem{armeni20163d}
Iro Armeni, Ozan Sener, Amir~R Zamir, Helen Jiang, Ioannis Brilakis, Martin
  Fischer, and Silvio Savarese.
\newblock {3D} semantic parsing of large-scale indoor spaces.
\newblock In {\em CVPR}, 2016.

\bibitem{Azuma2021ScanQA3Q}
Daich Azuma, Taiki Miyanishi, Shuhei Kurita, and Motoki Kawanabe.
\newblock {ScanQA}: 3d question answering for spatial scene understanding.
\newblock {\em ArXiv}, abs/2112.10482, 2021.

\bibitem{Bakr2022LookAA}
Eslam~Mohamed Bakr, Yasmeen Alsaedy, and Mohamed Elhoseiny.
\newblock Look around and refer: 2d synthetic semantics knowledge distillation
  for 3d visual grounding.
\newblock {\em ArXiv}, abs/2211.14241, 2022.

\bibitem{Banerjee2005METEORAA}
Satanjeev Banerjee and Alon Lavie.
\newblock Meteor: An automatic metric for mt evaluation with improved
  correlation with human judgments.
\newblock In {\em IEEvaluation@ACL}, 2005.

\bibitem{cai20223djcg}
Daigang Cai, Lichen Zhao, Jing Zhang, Lu Sheng, and Dong Xu.
\newblock 3djcg: A unified framework for joint dense captioning and visual
  grounding on 3d point clouds.
\newblock In {\em Proceedings of the IEEE/CVF Conference on Computer Vision and
  Pattern Recognition}, pages 16464--16473, 2022.

\bibitem{angel_phd_thesis}
Angel~X. Chang.
\newblock {\em Text to 3D Scene Generation}.
\newblock PhD thesis, Stanford University, 2015.

\bibitem{shapenet}
Angel~X. Chang, Thomas~A. Funkhouser, Leonidas~J. Guibas, Pat Hanrahan,
  Qi{-}Xing Huang, Zimo Li, Silvio Savarese, Manolis Savva, Shuran Song, Hao
  Su, Jianxiong Xiao, Li Yi, and Fisher Yu.
\newblock {ShapeNet}: An information-rich {3D} model repository.
\newblock {\em Computing Research Repository (CoRR)}, abs/1512.03012, 2015.

\bibitem{chen2022unit3d}
Dave~Zhenyu Chen, Ronghang Hu, Xinlei Chen, Matthias Nießner, and Angel~X.
  Chang.
\newblock Unit3d: A unified transformer for 3d dense captioning and visual
  grounding, 2022.

\bibitem{chen2021d3net}
Dave~Zhenyu Chen, Qirui Wu, Matthias Nießner, and Angel~X. Chang.
\newblock D3net: A speaker-listener architecture for semi-supervised dense
  captioning and visual grounding in rgb-d scans, 2021.

\bibitem{Chen2017LearningEO}
Guobin Chen, Wongun Choi, Xiang Yu, Tony~X. Han, and Manmohan Chandraker.
\newblock Learning efficient object detection models with knowledge
  distillation.
\newblock In {\em NIPS}, 2017.

\bibitem{chen2018text2shape}
Kevin Chen, Christopher~B Choy, Manolis Savva, Angel~X Chang, Thomas
  Funkhouser, and Silvio Savarese.
\newblock Text2shape: Generating shapes from natural language by learning joint
  embeddings.
\newblock {\em Computing Research Repository (CoRR)}, abs/1803.08495, 2018.

\bibitem{chen2021scan2cap}
Zhenyu Chen, Ali Gholami, Matthias Nie{\ss}ner, and Angel~X Chang.
\newblock {Scan2Cap}: Context-aware dense captioning in rgb-d scans.
\newblock In {\em Proceedings of the IEEE/CVF Conference on Computer Vision and
  Pattern Recognition}, pages 3193--3203, 2021.

\bibitem{zhenyu2019scanrefer}
Z.~Dave Chen, Angel~X. Chang, and Matthias Nie{\ss}ner.
\newblock {ScanRefer}: {3D} object localization in {RGB-D} scans using natural
  language.
\newblock {\em Computing Research Repository (CoRR)}, abs/1912.08830, 2019.

\bibitem{10.1007/978-3-642-35142-6_14}
Kevin Crowston.
\newblock Amazon mechanical turk: A research tool for organizations and
  information systems scholars.
\newblock In Anol Bhattacherjee and Brian Fitzgerald, editors, {\em Shaping the
  Future of ICT Research. Methods and Approaches}, pages 210--221, Berlin,
  Heidelberg, 2012. Springer Berlin Heidelberg.

\bibitem{ScanNet}
Angela Dai, Angel~X. Chang, Manolis Savva, Maciej Halber, Thomas Funkhouser,
  and Nie{\ss}ner.
\newblock {ScanNet}: Richly-annotated {3D} reconstructions of indoor scenes.
\newblock In {\em CVPR}, 2017.

\bibitem{graph2scene2021}
Helisa Dhamo, Fabian Manhardt, Nassir Navab, and Federico Tombari.
\newblock Graph-to-3d: End-to-end generation and manipulation of 3d scenes
  using scene graphs.
\newblock In {\em IEEE International Conference on Computer Vision (ICCV)},
  2021.

\bibitem{Feng2021FreeformDG}
Mingtao Feng, Zhen Li, Qi Li, Liang Zhang, Xiangdong Zhang, Guangming Zhu, Hui
  Zhang, Yaonan Wang, and Ajmal~S. Mian.
\newblock Free-form description guided 3d visual graph network for object
  grounding in point cloud.
\newblock {\em 2021 IEEE/CVF International Conference on Computer Vision
  (ICCV)}, pages 3702--3711, 2021.

\bibitem{fu20213d}
Huan Fu, Bowen Cai, Lin Gao, Ling-Xiao Zhang, Jiaming Wang, Cao Li, Qixun Zeng,
  Chengyue Sun, Rongfei Jia, Binqiang Zhao, et~al.
\newblock 3d-front: 3d furnished rooms with layouts and semantics.
\newblock In {\em Proceedings of the IEEE/CVF International Conference on
  Computer Vision}, pages 10933--10942, 2021.

\bibitem{gordon2018iqa}
Daniel Gordon, Aniruddha Kembhavi, Mohammad Rastegari, Joseph Redmon, Dieter
  Fox, and Ali Farhadi.
\newblock Iqa: Visual question answering in interactive environments.
\newblock In {\em Proceedings of the IEEE conference on computer vision and
  pattern recognition}, pages 4089--4098, 2018.

\bibitem{lstm}
Klaus Greff, Rupesh~Kumar Srivastava, Jan Koutn{\'i}k, Bas~R. Steunebrink, and
  J{\"u}rgen Schmidhuber.
\newblock Lstm: A search space odyssey.
\newblock {\em IEEE Transactions on Neural Networks and Learning Systems},
  28:2222--2232, 2017.

\bibitem{han2020shapecaptioner}
Zhizhong Han, Chao Chen, Yu-Shen Liu, and Matthias Zwicker.
\newblock Shapecaptioner: Generative caption network for 3d shapes by learning
  a mapping from parts detected in multiple views to sentences.
\newblock In {\em ACM International Conference on Multimedia (MM)}, 2020.

\bibitem{TransRefer3D}
Dailan He, Yusheng Zhao, Junyu Luo, Tianrui Hui, Shaofei Huang, Aixi Zhang, and
  Si Liu.
\newblock {TransRefer3D}: Entity-and-relation aware transformer for
  fine-grained {3D} visual grounding.
\newblock {\em Computing Research Repository (CoRR)}, abs/2108.02388, 2021.

\bibitem{hong2021vlgrammar}
Yining Hong, Qing Li, Song-Chun Zhu, and Siyuan Huang.
\newblock {VLGrammar}: Grounded grammar induction of vision and language.
\newblock {\em ICCV}, 2021.

\bibitem{hou2021exploring}
Ji Hou, Benjamin Graham, Matthias Nie{\ss}ner, and Saining Xie.
\newblock Exploring data-efficient 3d scene understanding with contrastive
  scene contexts.
\newblock In {\em Proceedings of the IEEE/CVF Conference on Computer Vision and
  Pattern Recognition}, pages 15587--15597, 2021.

\bibitem{hua2016scenenn}
Binh-Son Hua, Quang-Hieu Pham, Duc~Thanh Nguyen, Minh-Khoi Tran, Lap-Fai Yu,
  and Sai-Kit Yeung.
\newblock Scenenn: A scene meshes dataset with annotations.
\newblock In {\em 2016 fourth international conference on 3D vision (3DV)},
  pages 92--101. Ieee, 2016.

\bibitem{LADIS}
Ian Huang, Panos Achlioptas, Tianyi Zhang, Sergey Tulyakov, Minhyuk Sung, and
  Guibas Leonidas.
\newblock {LADIS}: Language disentanglement for {3D} shape editing.
\newblock In {\em Findings of Empirical Methods in Natural Language
  Processing}, 2022.

\bibitem{huang2022multi}
Shijia Huang, Yilun Chen, Jiaya Jia, and Liwei Wang.
\newblock Multi-view transformer for 3d visual grounding.
\newblock In {\em CVPR}, 2022.

\bibitem{wald}
Wald Johanna, Avetisyan Armen, Navab Nassir, Tombari Federico, and Niessner
  Matthias.
\newblock Rio: 3d object instance re-localization in changing indoor
  environments.
\newblock {\em Proceedings IEEE International Conference on Computer Vision
  (ICCV)}, 2019.

\bibitem{kazemzadeh2014referitgame}
Sahar Kazemzadeh, Vicente Ordonez, Mark Matten, and Tamara Berg.
\newblock Referitgame: Referring to objects in photographs of natural scenes.
\newblock In {\em Proceedings of the 2014 conference on empirical methods in
  natural language processing (EMNLP)}, pages 787--798, 2014.

\bibitem{kolve2017ai2}
Eric Kolve, Roozbeh Mottaghi, Winson Han, Eli VanderBilt, Luca Weihs, Alvaro
  Herrasti, Daniel Gordon, Yuke Zhu, Abhinav Gupta, and Ali Farhadi.
\newblock Ai2-thor: An interactive 3d environment for visual ai.
\newblock {\em arXiv preprint arXiv:1712.05474}, 2017.

\bibitem{PartGlot}
Juil Koo, Ian Huang, Panos Achlioptas, Leonidas~J. Guibas, and Minhyuk Sung.
\newblock {PartGlot}: Learning shape part segmentation from language reference
  games.
\newblock In {\em CVPR}, 2022.

\bibitem{Li20223DCC}
Yuchen Li, Ujjwal Upadhyay, Habib Slim, Ahmed Abdelreheem, Arpita Prajapati,
  Suhail Pothigara, Peter Wonka, and Mohamed Elhoseiny.
\newblock 3d compat: Composition of materials on parts of 3d things.
\newblock In {\em European Conference on Computer Vision}, 2022.

\bibitem{Lin2004ROUGEAP}
Chin-Yew Lin.
\newblock Rouge: A package for automatic evaluation of summaries.
\newblock In {\em ACL 2004}, 2004.

\bibitem{luo20223d}
Junyu Luo, Jiahui Fu, Xianghao Kong, Chen Gao, Haibing Ren, Hao Shen, Huaxia
  Xia, and Si Liu.
\newblock 3d-sps: Single-stage 3d visual grounding via referred point
  progressive selection.
\newblock {\em arXiv preprint arXiv:2204.06272}, 2022.

\bibitem{ma2023sqa3d}
Xiaojian Ma, Silong Yong, Zilong Zheng, Qing Li, Yitao Liang, Song-Chun Zhu,
  and Siyuan Huang.
\newblock Sqa3d: Situated question answering in 3d scenes, 2023.

\bibitem{mao2016generation}
Junhua Mao, Jonathan Huang, Alexander Toshev, Oana Camburu, Alan~L Yuille, and
  Kevin Murphy.
\newblock Generation and comprehension of unambiguous object descriptions.
\newblock In {\em Proceedings of the IEEE conference on computer vision and
  pattern recognition}, pages 11--20, 2016.

\bibitem{nichols-botros-2015-sprl}
Eric Nichols and Fadi Botros.
\newblock {S}p{RL}-{CWW}: Spatial relation classification with independent
  multi-class models.
\newblock In {\em Proceedings of the 9th International Workshop on Semantic
  Evaluation ({S}em{E}val 2015)}. Association for Computational Linguistics,
  2015.

\bibitem{Papineni2002BleuAM}
Kishore Papineni, Salim Roukos, Todd Ward, and Wei-Jing Zhu.
\newblock Bleu: a method for automatic evaluation of machine translation.
\newblock In {\em ACL}, 2002.

\bibitem{plummer2015flickr30k}
Bryan~A Plummer, Liwei Wang, Chris~M Cervantes, Juan~C Caicedo, Julia
  Hockenmaier, and Svetlana Lazebnik.
\newblock Flickr30k entities: Collecting region-to-phrase correspondences for
  richer image-to-sentence models.
\newblock In {\em CVPR}, 2015.

\bibitem{DBLP:journals/corr/abs-2010-16061}
David M.~W. Powers.
\newblock Evaluation: from precision, recall and f-measure to roc,
  informedness, markedness and correlation.
\newblock {\em CoRR}, abs/2010.16061, 2020.

\bibitem{qi2017pointnetpp}
Charles~Ruizhongtai Qi, Li Yi, Hao Su, and Leonidas~J Guibas.
\newblock {PointNet++}: Deep hierarchical feature learning on point sets in a
  metric space.
\newblock In {\em NeurIPS}, 2017.

\bibitem{reizenstein2021common}
Jeremy Reizenstein, Roman Shapovalov, Philipp Henzler, Luca Sbordone, Patrick
  Labatut, and David Novotny.
\newblock Common objects in 3d: Large-scale learning and evaluation of
  real-life 3d category reconstruction.
\newblock In {\em Proceedings of the IEEE/CVF International Conference on
  Computer Vision}, pages 10901--10911, 2021.

\bibitem{LanguageRefer}
Junha Roh, Karthik Desingh, Ali Farhadi, and Dieter Fox.
\newblock {LanguageRefer}: Spatial-language model for {3D} visual grounding.
\newblock {\em Computing Research Repository (CoRR)}, abs/2107.03438, 2021.

\bibitem{Rozenberszki2022LanguageGroundedI3}
D{\'a}vid Rozenberszki, Or Litany, and Angela Dai.
\newblock Language-grounded indoor 3d semantic segmentation in the wild.
\newblock {\em ArXiv}, abs/2204.07761, 2022.

\bibitem{sharma2022denserefer3d}
Akshit Sharma.
\newblock Denserefer3d: A language and 3d dataset for coreference resolution
  and referring expression comprehension, 2022.

\bibitem{silberman2012indoor}
Nathan Silberman, Derek Hoiem, Pushmeet Kohli, and Rob Fergus.
\newblock Indoor segmentation and support inference from rgbd images.
\newblock In {\em European conference on computer vision}, pages 746--760.
  Springer, 2012.

\bibitem{SNARE}
Jesse Thomason, Mohit Shridhar, Yonatan Bisk, Chris Paxton, and Luke
  Zettlemoyer.
\newblock Language grounding with 3d objects.
\newblock {\em Computing Research Repository (CoRR)}, abs/2107.12514, 2021.

\bibitem{Vedantam2015CIDErCI}
Ramakrishna Vedantam, C.~Lawrence Zitnick, and Devi Parikh.
\newblock Cider: Consensus-based image description evaluation.
\newblock {\em 2015 IEEE Conference on Computer Vision and Pattern Recognition
  (CVPR)}, pages 4566--4575, 2015.

\bibitem{vinyals2015show}
Oriol Vinyals, Alexander Toshev, Samy Bengio, and Dumitru Erhan.
\newblock Show and tell: A neural image caption generator.
\newblock In {\em CVPR}, 2015.

\bibitem{wang2023text}
Guangzhi Wang, Hehe Fan, and Mohan Kankanhalli.
\newblock Text to point cloud localization with relation-enhanced transformer.
\newblock {\em arXiv preprint arXiv:2301.05372}, 2023.

\bibitem{Wang2022SpatialityguidedTF}
Heng Wang, Chaoyi Zhang, Jianhui Yu, and Weidong~(Tom) Cai.
\newblock Spatiality-guided transformer for 3d dense captioning on point
  clouds.
\newblock In {\em IJCAI}, 2022.

\bibitem{wijmans2019embodied}
Erik Wijmans, Samyak Datta, Oleksandr Maksymets, Abhishek Das, Georgia
  Gkioxari, Stefan Lee, Irfan Essa, Devi Parikh, and Dhruv Batra.
\newblock Embodied question answering in photorealistic environments with point
  cloud perception.
\newblock In {\em Proceedings of the IEEE/CVF Conference on Computer Vision and
  Pattern Recognition}, pages 6659--6668, 2019.

\bibitem{teacher_forcing}
Ronald~J. Williams and David Zipser.
\newblock A learning algorithm for continually running fully recurrent neural
  networks.
\newblock {\em Neural Computation}, 1989.

\bibitem{wu2022eda}
Yanmin Wu, Xinhua Cheng, Renrui Zhang, Zesen Cheng, and Jian Zhang.
\newblock Eda: Explicit text-decoupling and dense alignment for 3d visual
  grounding, 2022.

\bibitem{xiao2013sun3d}
Jianxiong Xiao, Andrew Owens, and Antonio Torralba.
\newblock Sun3d: A database of big spaces reconstructed using sfm and object
  labels.
\newblock In {\em Proceedings of the IEEE international conference on computer
  vision}, pages 1625--1632, 2013.

\bibitem{xu2015show}
Kelvin Xu, Jimmy Ba, Ryan Kiros, Aaron Courville, Ruslan Salakhutdinov, Richard
  Zemel, and Yoshua Bengio.
\newblock Show, attend and tell: Neural image caption generation with visual
  attention.
\newblock In {\em International Conference on Machine Learning (ICML)}, 2015.

\bibitem{Yang_2020_CVPR}
Muli Yang, Cheng Deng, Junchi Yan, Xianglong Liu, and Dacheng Tao.
\newblock Learning unseen concepts via hierarchical decomposition and
  composition.
\newblock In {\em CVPR}, 2020.

\bibitem{yang2020improving}
Zhengyuan Yang, Tianlang Chen, Liwei Wang, and Jiebo Luo.
\newblock Improving one-stage visual grounding by recursive sub-query
  construction.
\newblock In {\em European Conference on Computer Vision}, pages 387--404.
  Springer, 2020.

\bibitem{yang2019fast}
Zhengyuan Yang, Boqing Gong, Liwei Wang, Wenbing Huang, Dong Yu, and Jiebo Luo.
\newblock A fast and accurate one-stage approach to visual grounding.
\newblock In {\em Proceedings of the IEEE/CVF International Conference on
  Computer Vision}, pages 4683--4693, 2019.

\bibitem{Yang2019AFA}
Zhengyuan Yang, Boqing Gong, Liwei Wang, Wenbing Huang, Dong Yu, and Jiebo Luo.
\newblock A fast and accurate one-stage approach to visual grounding.
\newblock {\em 2019 IEEE/CVF International Conference on Computer Vision
  (ICCV)}, pages 4682--4692, 2019.

\bibitem{SAT_3D}
Zhengyuan Yang, Songyang Zhang, Liwei Wang, and Jiebo Luo.
\newblock {SAT:} 2d semantics assisted training for {3D} visual grounding.
\newblock {\em Computing Research Repository (CoRR)}, abs/2105.11450, 2021.

\bibitem{yu2019multi}
Licheng Yu, Xinlei Chen, Georgia Gkioxari, Mohit Bansal, Tamara~L Berg, and
  Dhruv Batra.
\newblock Multi-target embodied question answering.
\newblock In {\em Proceedings of the IEEE/CVF Conference on Computer Vision and
  Pattern Recognition}, pages 6309--6318, 2019.

\bibitem{yu2018mattnet}
Licheng Yu, Zhe Lin, Xiaohui Shen, Jimei Yang, Xin Lu, Mohit Bansal, and
  Tamara~L Berg.
\newblock Mattnet: Modular attention network for referring expression
  comprehension.
\newblock In {\em Proceedings of the IEEE Conference on Computer Vision and
  Pattern Recognition}, pages 1307--1315, 2018.

\bibitem{yu2016modeling}
Licheng Yu, Patrick Poirson, Shan Yang, Alexander~C Berg, and Tamara~L Berg.
\newblock Modeling context in referring expressions.
\newblock In {\em European Conference on Computer Vision}, pages 69--85.
  Springer, 2016.

\bibitem{yuan2022toward}
Zhihao Yuan, Xu Yan, Zhuo Li, Xuhao Li, Yao Guo, Shuguang Cui, and Zhen Li.
\newblock Toward explainable and fine-grained 3d grounding through referring
  textual phrases.
\newblock {\em arXiv preprint arXiv:2207.01821}, 2022.

\bibitem{Yuan_2022_CVPR}
Zhihao Yuan, Xu Yan, Yinghong Liao, Yao Guo, Guanbin Li, Shuguang Cui, and Zhen
  Li.
\newblock X-trans2cap: Cross-modal knowledge transfer using transformer for 3d
  dense captioning.
\newblock In {\em Proceedings of the IEEE/CVF Conference on Computer Vision and
  Pattern Recognition (CVPR)}, pages 8563--8573, June 2022.

\bibitem{InstanceRefer}
Zhihao Yuan, Xu Yan, Yinghong Liao, Ruimao Zhang, Zhen Li, and Shuguang Cui.
\newblock {InstanceRefer}: Cooperative holistic understanding for visual
  grounding on point clouds through instance multi-level contextual referring.
\newblock {\em ICCV}, 2021.

\bibitem{zhao2021_3DVG_Transformer}
Lichen Zhao, Daigang Cai, Lu Sheng, and Dong Xu.
\newblock {3DVG-Transformer}: Relation modeling for visual grounding on point
  clouds.
\newblock In {\em ICCV}, 2021.

\bibitem{zhong2022contextual}
Yufeng Zhong, Long Xu, Jiebo Luo, and Lin Ma.
\newblock Contextual modeling for 3d dense captioning on point clouds, 2022.

\end{thebibliography}
}

%%%%%%%%% SUPPLEMENTARY TEXT
% \clearpage
\appendix
\begin{figure*}[!htb]
    \includegraphics[width=0.7\linewidth]{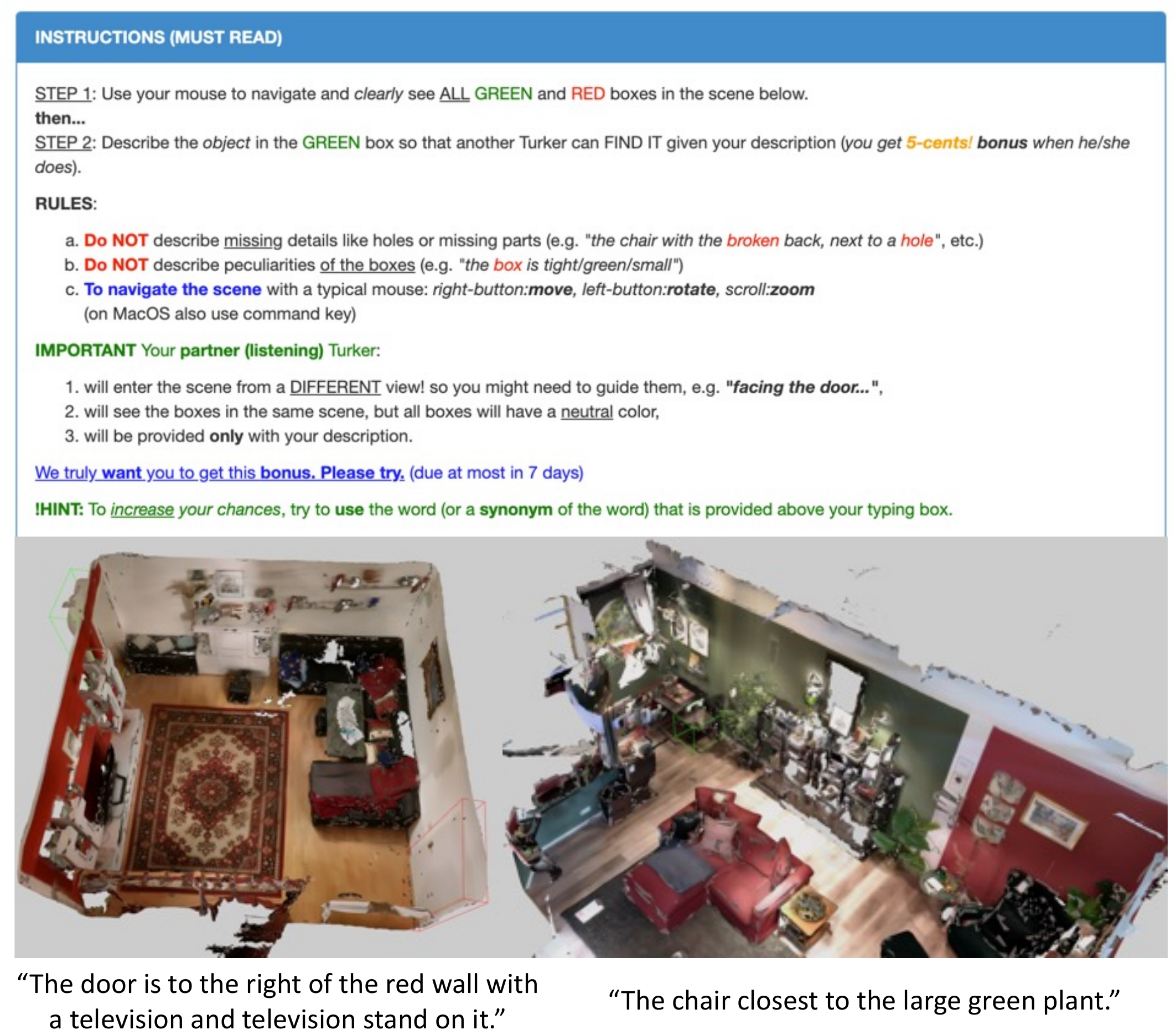}
    \caption{\textbf{User interface for the collection of referential sentences for the 3RScan zero-shot experiment}. On the top, we show the detailed instructions provided to the annotator to ensure the task requirements are clear and straightforward. On the bottom (b), we show two examples of the resulting annotations. The target objects are the ones inside the green bounding boxes, while the same class distractor objects are in the red bounding boxes.}
    \label{fig:trscanAMT}
\end{figure*}

\begin{figure*}[!htb]%
  \centering
    \includegraphics[width=0.8\linewidth]{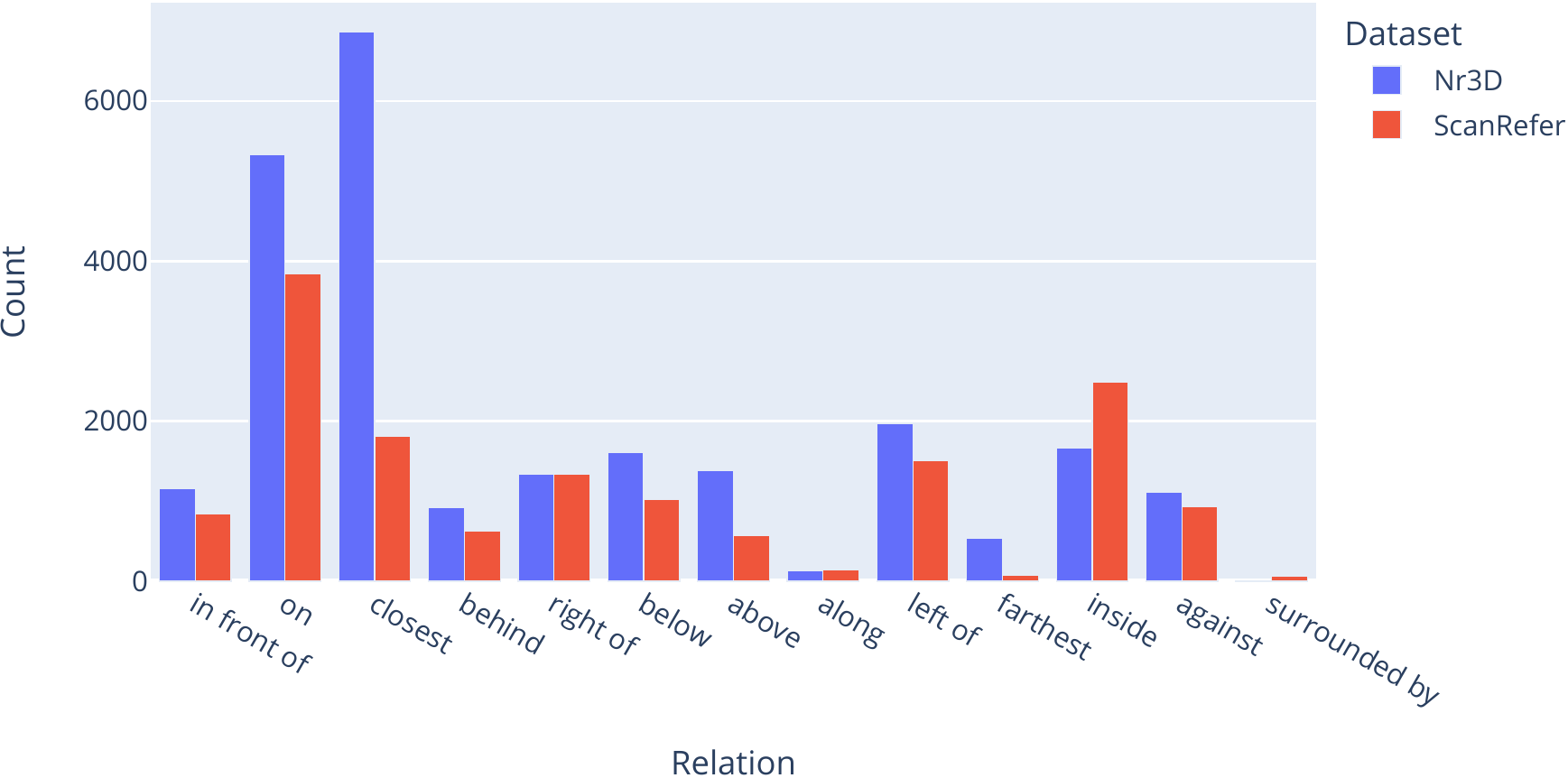} 
    \caption{\textbf{Breakdown of the extracted pairwise spatial relationships of \nr{} and \scanRefer{} datasets.} Despite their similar nature, we see that in terms of spatial relations types used to describe objects, there is a noticeable discrepancy among their annotations.}
    \label{fig:spa_rel}
\end{figure*}
\begin{figure*}[!htb]%
    \centering
    \subfloat[]{{\includegraphics[width=7cm]{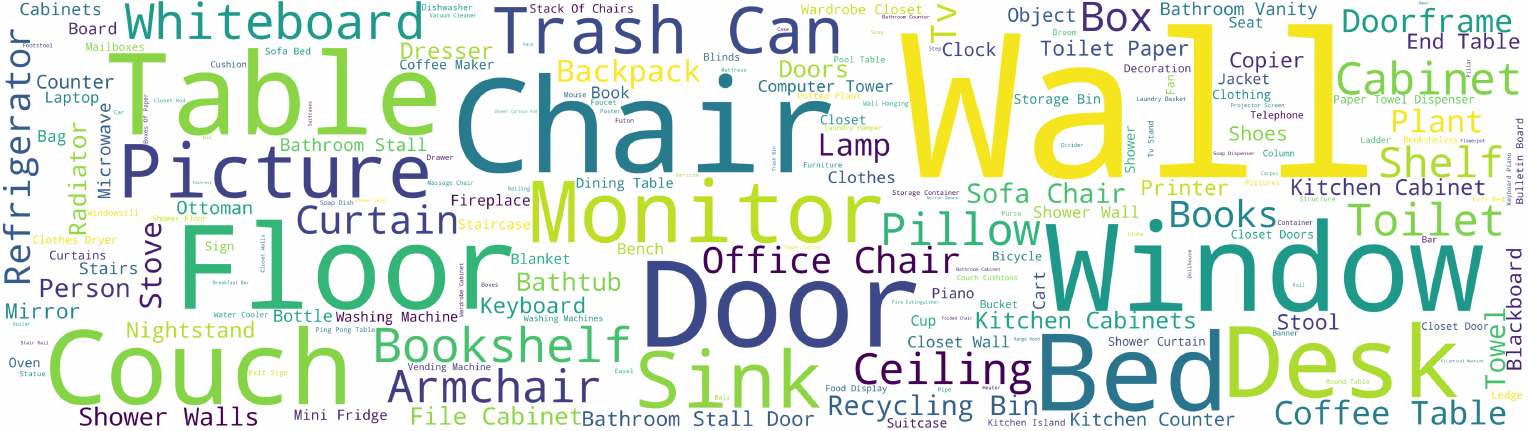} }}%
    \qquad
    \subfloat[\centering]{{\includegraphics[width=7cm]{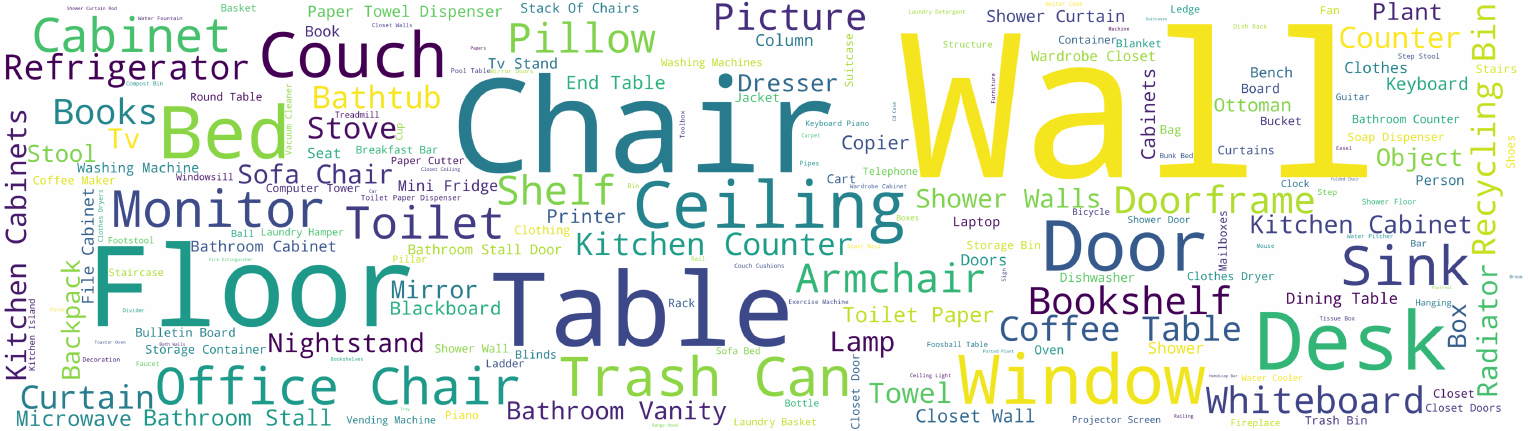} }}
    \caption{\textbf{Wordclouds depicting the most common object classes in (a) \nr{} and (b) \scanRefer{} datasets.} The font size of each printed class name is proportional to its underlying frequency (better seen by zooming in).}
    \label{fig:anchor_images}
\end{figure*}

\begin{figure*}[!htb]%
  \centering
    \subfloat[]{{\includegraphics[width=0.8\linewidth]{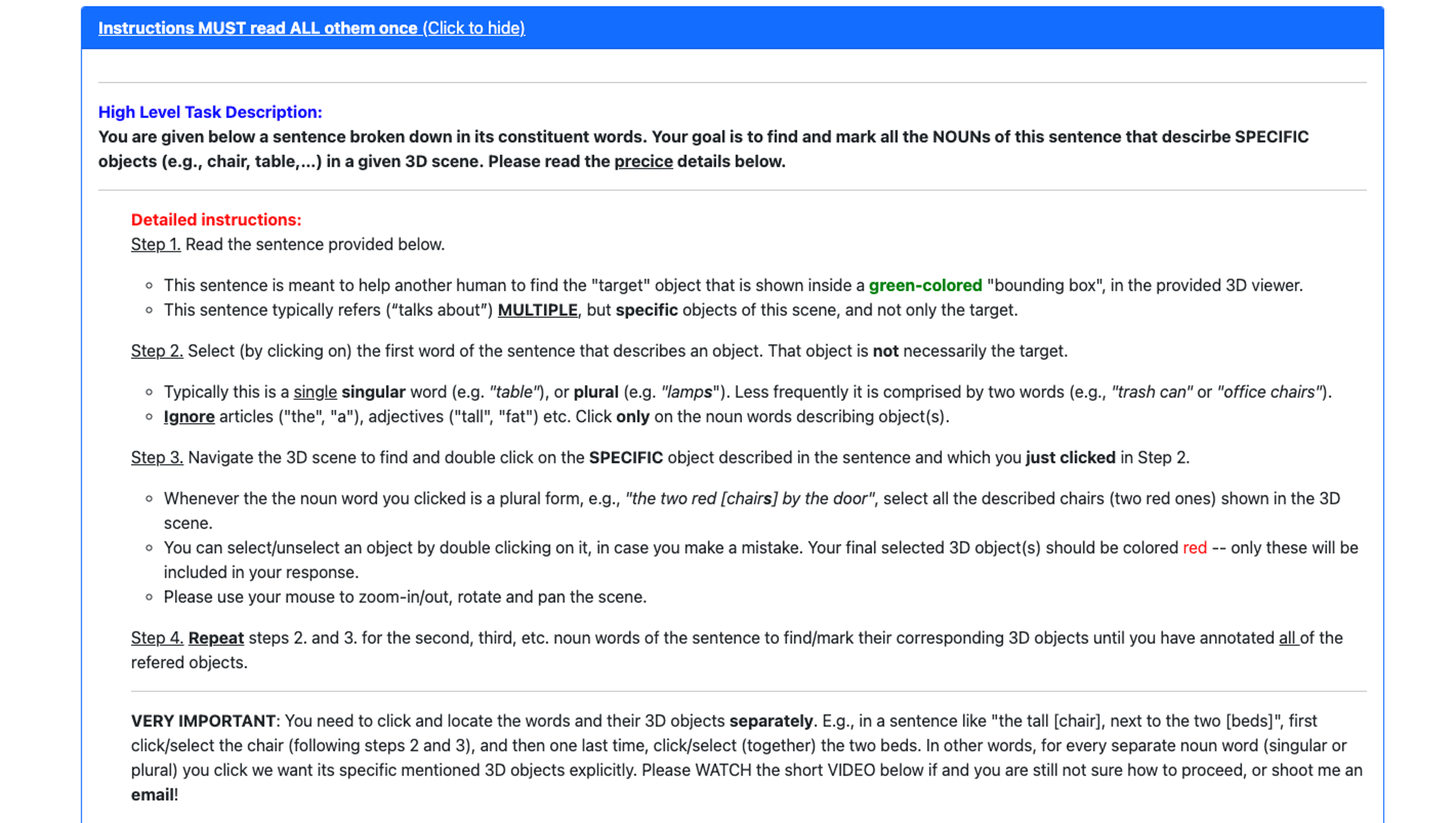} }}%
    \qquad
    \subfloat[\centering]{{\includegraphics[width=0.8\linewidth]{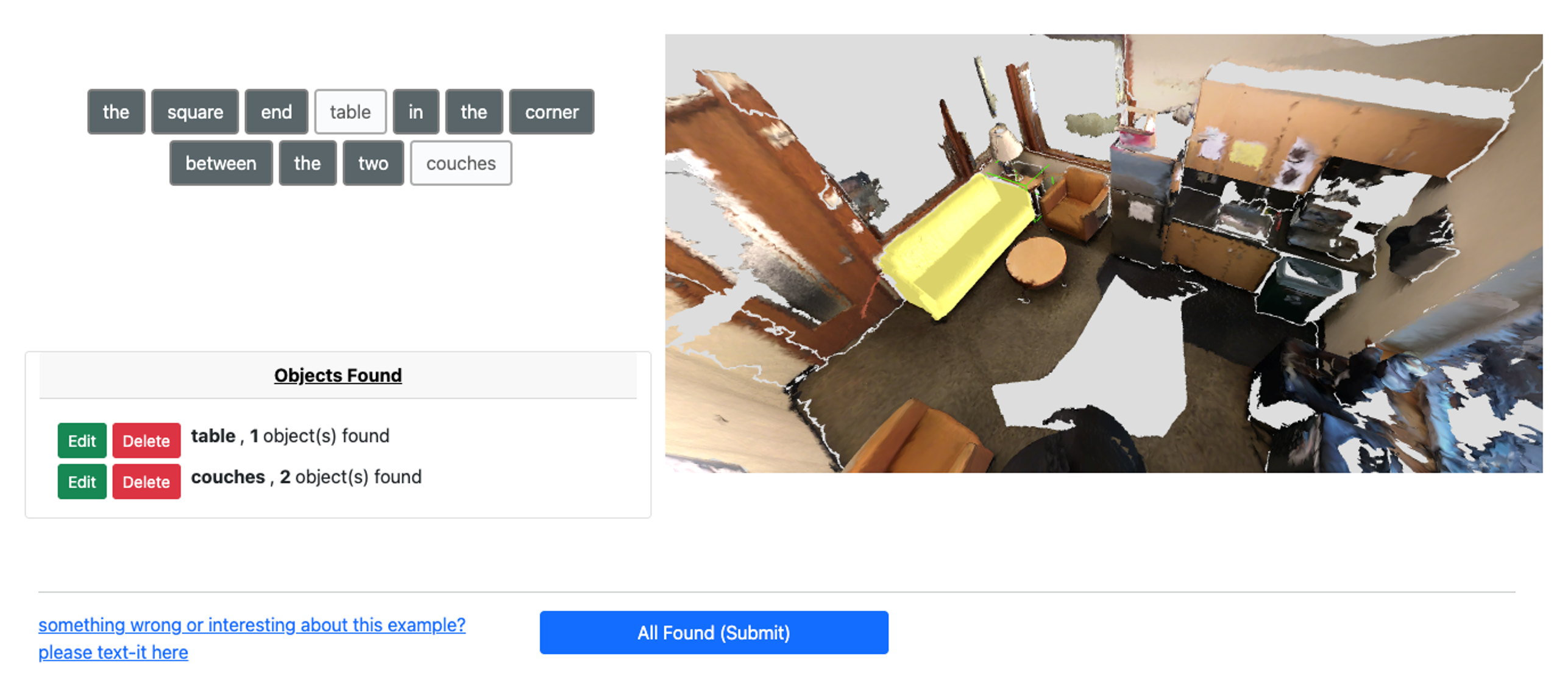} }}
    \caption{\textbf{User interface for the \datasetName{} dataset collection}. On the top (a), we show the detailed instructions provided to the annotator to ensure the task requirements are clear and straightforward. On the bottom (b), we show an example of a resulting annotation.}
    \label{fig:ui}
\end{figure*}

\section{Zero-Shot Experiments}

In this section, we discuss our zero-shot experiments on the 3RScan dataset~\cite{wald}. First, we discuss the collection of referential sentences. 3RScan is a large-scale, real-world dataset that contains 1482 3D reconstructions. Second, we report the zero-shot listening accuracy of our proposed model MVT-\datasetSuffix{} compared to the original MVT model.

\subsection{Referential Sentences Collection for 3RScan}
We collect referential sentences for the validation scans present in the 3RScan dataset. We follow the data collection approach presented in~\cite{achlioptas2020referit_3d}. The dataset collection pipeline consists of two stages; data collection and data verification. In \Cref{fig:trscanAMT}, we show the AMT interface used for data collection along with actual collected data examples. We collect in total 840 referential sentences covering all of the 47 scans of the official validation split. 

\subsection{Zero-Shot Listening Results}
We do zero-shot neural listening tests using a pre-trained MVT-\datasetSuffix{} model, which is trained on Nr3D using the rich annotations of ScanEnts3D and our novel proposed losses and using an original MVT model trained on Nr3D as in~\cite{huang2022multi} without \datasetName{}. We center the input scene point cloud around the origin point and transform the point cloud to become axis-aligned as described in~\cite{graph2scene2021}. In \Cref{tab:zs}, MVT-\datasetSuffix{} outperforms MVT on out-of-domain 3D scenes by 4.17\%. The result shows that neural listeners when trained on \datasetName{}, can exhibit better 3D scene understanding even on unseen scans.

\begin{table}[!htb]
    \centering
    \begin{tabular}{lc}
    Method & Overall Acc. \\
    \toprule
    MVT~\cite{huang2022multi}  & 11.80\%  \\
    MVT-\datasetSuffix{} &  \textbf{15.97\%} \textbf{\textcolor[RGB]{102,178,84}{(+4.17\%)}} \\
    \bottomrule
    \end{tabular}
    \caption{Zero-Shot listening performance on our collected referential sentences on the 3RScan dataset. MVT-\datasetSuffix{} outperforms the original MVT model on test examples of unseen scans.}
    \label{tab:zs}
\end{table}

\section{\datasetName{} Dataset Analysis}
In this section, we provide a more detailed analysis of our proposed \datasetName{} dataset. In \Cref{fig:spa_rel}, we show a breakdown of the extracted pairwise spatial relationships between the scan entities in \datasetName{}. In total, we extract using existing spatial relation classifiers~\cite{nichols-botros-2015-sprl} 24,028 pairwise spatial relations for the \nr{} dataset and 15,278 pairwise spatial relations for the \scanRefer{} dataset. 

In \Cref{fig:anchor_images}, we show the classes most used as anchor objects for both \nr{} and \scanRefer{} datasets. We observe that the most used anchor classes are walls, chairs, windows, and doors. We also observe that only 363 fine-grained object classes are used for the anchor objects.

In \Cref{fig:n_scan_entities}, we show a histogram of the number of scan entities of \datasetName{} for the \nr{} and \scanRefer{} datasets. The mean number of scan entities in \nr{} is 2.5, with a standard deviation of 1.17. The mean number of scan entities in \scanRefer{} is 3.96 with a standard deviation of 1.45.

\begin{minipage}{\linewidth}
\begin{minipage}[t]{0.45\linewidth}
\centering
\includegraphics[width=1.0\linewidth]{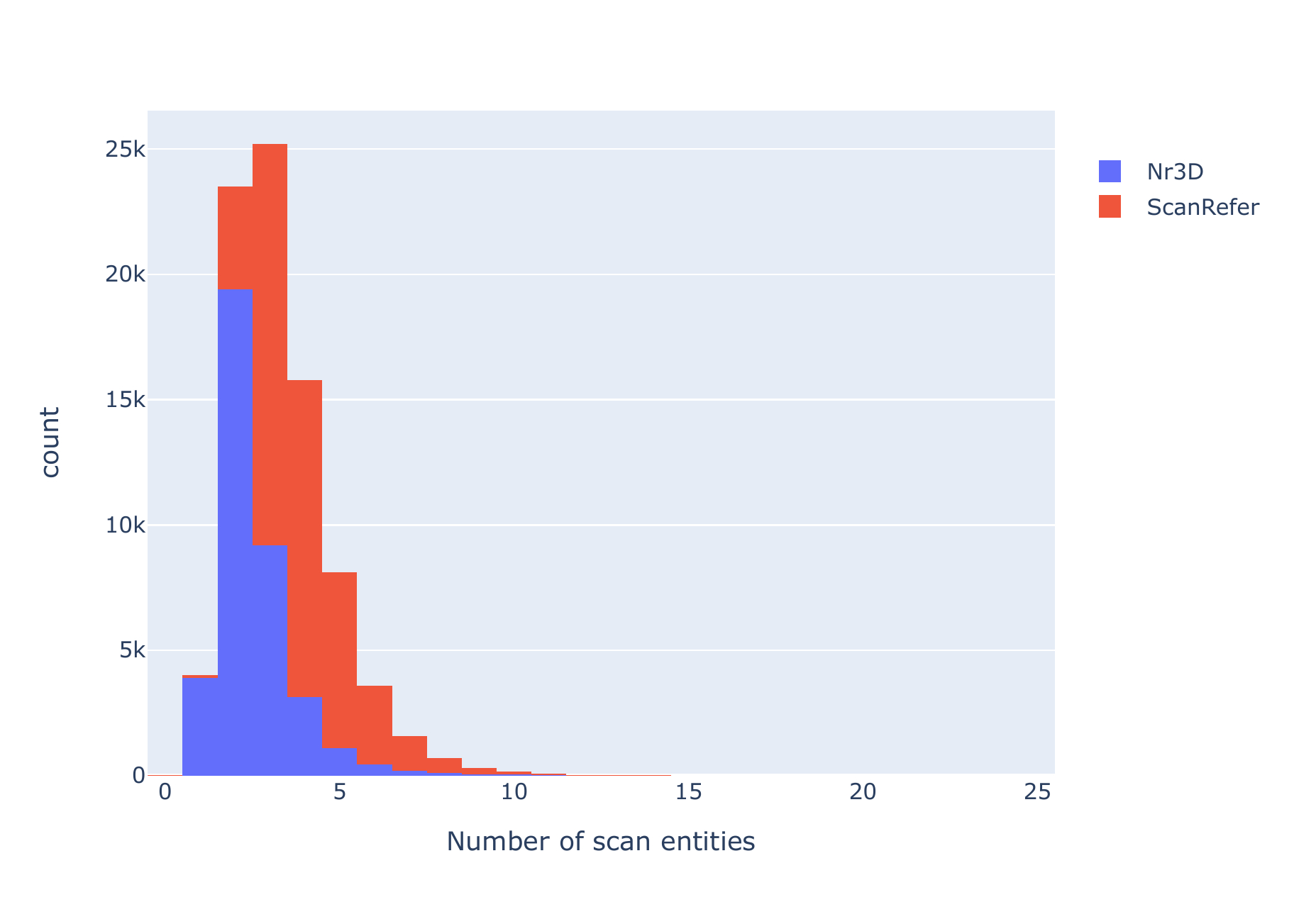}
\captionsetup{type=figure}
\captionof{figure}{\textbf{Histogram of the number of \datasetName{} scan entities present in \nr{} and \scanRefer{} datasets}.}
\label{fig:n_scan_entities}
\end{minipage}
\hfill
\begin{minipage}[t]{0.5\linewidth}
\centering
\includegraphics[width=1.0\linewidth]{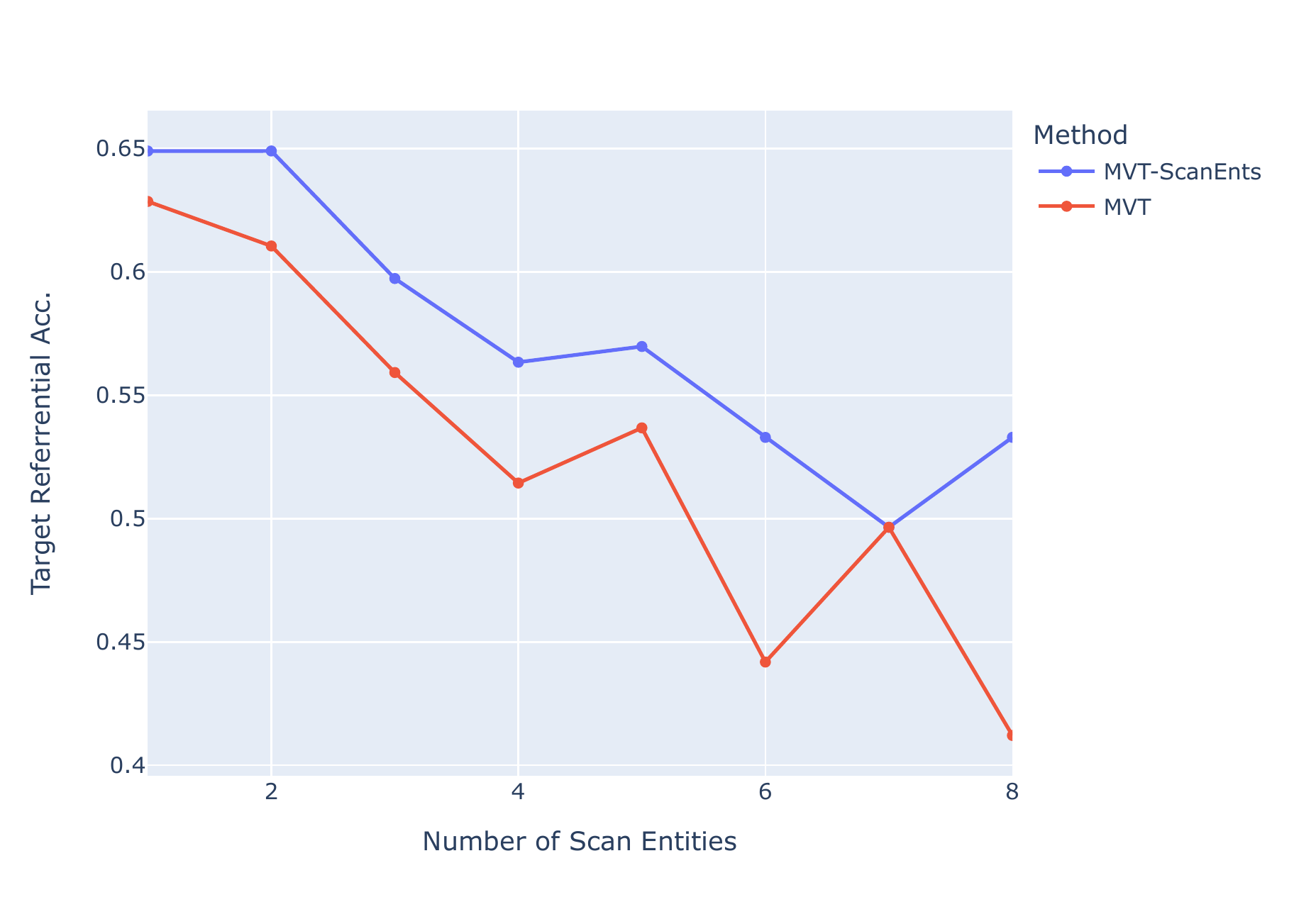}
\captionsetup{type=figure}
\captionof{figure}{\textbf{Comparison between the performance of MVT-\datasetSuffix{} and MVT models when increasing the number of scan entities and the number of same-class distractor objects.} The performance generally decreases when increasing the number of the scan entities and the same-class distractor objects.}
    \label{fig:comp_anchors}
\end{minipage}
\end{minipage}

\section{\datasetName{} Dataset Collection}
This section discusses in detail the two phases of our \datasetName{} curation. Figure \ref{fig:ui} shows the user interface we implemented.

\textbf{Annotation Phase.} An annotator is given an utterance and a 3D scene. While the utterance generally describes one specific object in the 3D scene, the annotator is first asked to mark all the nouns (entities) that describe specific objects in the given 3D scene (e.g., chair, table, etc.) in the utterance. Then, for each selected entity in the given utterance, the annotator must highlight the corresponding 3D objects in the given 3D scene. The annotator can zoom, pan or rotate the 3D scene to  find the corresponding 3D objects. Each annotator is provided with one random utterance at a time. We assign one annotator for each example.

\textbf{Review Phase.} A reviewer is given one annotated example randomly and is asked to determine whether the example was correctly annotated. If the example was annotated incorrectly, the reviewer is then requested to correct and fix the annotation. The reviewer is shown a similar user interface to the annotator. Each submission is reviewed by one reviewer.

\begin{figure*}[!htb]
    \centering
    \includegraphics[width=0.8\linewidth]{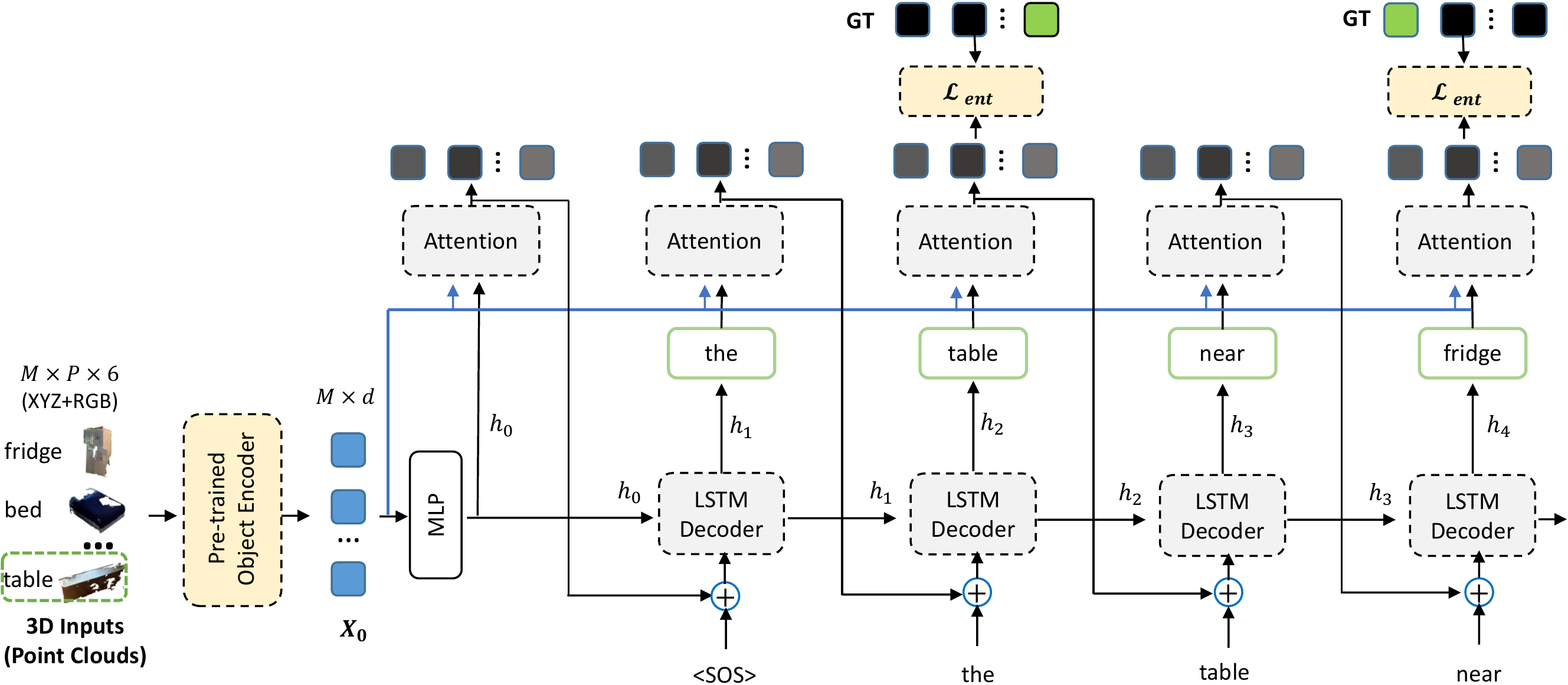}
    \caption{\textbf{Our proposed SATCap-\datasetSuffix{} model.} SATCap-\datasetSuffix{} is based on the ``Show, Attend, and Tell" model~\cite{vinyals2015show}. We use a pre-trained 3D object encoder for encoding the scene objects. The decoder is an LSTM~\cite{lstm}, where we apply our proposed loss $\mathcal{L}_{ent}$ during training. If the word to be predicted by the decoder in the current time-step (like table and fridge) corresponds to a scan entity in the target caption, the attention values to the 3D objects that belong to the scan entity should be higher than that of the objects that do not belong to that scene entity.}
    \label{fig:sat_cap_model}
\end{figure*}

\section{Neural Listeners}
\subsection{SAT-\datasetSuffix{}}
\begin{figure*}[!htb]
    \centering
    \includegraphics[width=0.7\linewidth]{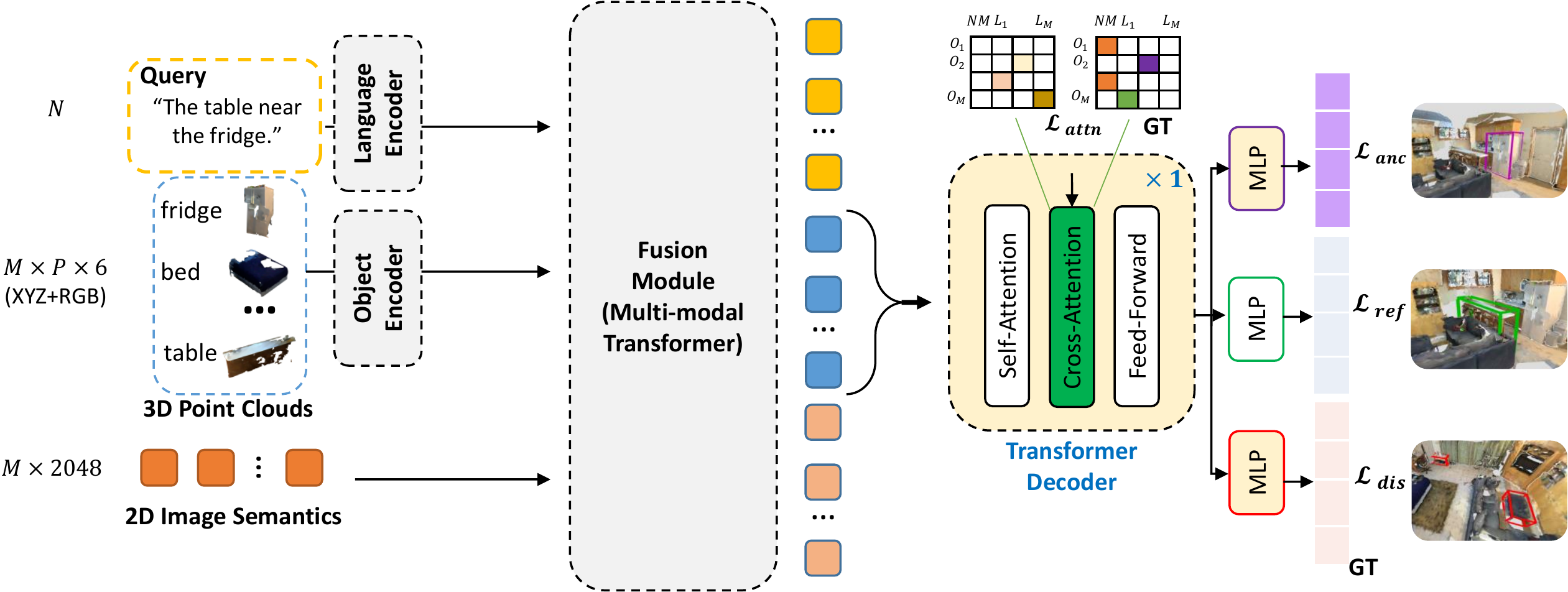}
    \caption{\textbf{Our proposed SAT-\datasetSuffix{} model.} We add a cross-attention layer operating on the 3D object and language features. Our proposed losses are applied after the added cross-attention layer in a similar manner to the MVT-
    datasetSuffix{} model.}
    \label{fig:sat_model}
\end{figure*}

This section discusses our modifications to the SAT~\cite{SAT_3D} neural listener. For the cross-attention map loss, since the SAT model is using transformer encoder layers, it does not contain an apparent cross-attention operation between the object tokens and the language tokens. To address that, we add one transformer decoder layer as shown in \Cref{fig:sat_model}, where we apply our proposed cross-attention map loss. The Cross-Attention Map loss encourages the network to attain high relevance values between both the objects and the words representing the same underlying scan entity. The target matrix $Y_\mathrm{attn}$ is a binary matrix of shape $M \times N$, where a cell ($y_{i, j}$) has a value of 1 if the $i$th object and the $j$th word correspond to one another. To cover the case of the 3D objects that do not belong to any of the scan entities in the given utterance, we add an extra word token called $<$NM$>$ as shown in \Cref{fig:sat_model} and for every object $k$ that does not belong to any of the scan entities, we set the cell ($y_{1, k}$ to the value of 1. The $<$NM$>$ mention token is always added after the $<$CLS$>$ token. The anchor prediction loss and the same-class distractor loss are applied to the late context-aware feature. 

\subsection{3DJCG-\datasetSuffix{}}
The 3DJCG~\cite{cai20223djcg} model is an object-detection-based model, where the input to the model is a point cloud of a 3D scene and an input utterance. The task of the model is to localize the target object via predicting an axis-aligned 3D bounding box around the target object. We apply the anchor prediction loss as discussed in the main paper in Section 4.1.1. We apply an MLP $\phi$ on the feature vectors of the detected object proposals to obtain a confidence score $x_i$ $\in [0, 1]$ of whether the object proposal is an anchor object or not.  To construct the ground truth for the anchor prediction loss, we follow a similar approach as in~\cite{Yang2019AFA, zhenyu2019scanrefer}. For each object proposal, the ground-truth label is $y_i$ $\in \{0, 1\}$. We set the label $y_j=1$ for the $j^{th}$ proposal that has the highest IOU with the box of one of the ground truth anchor objects. We apply a binary cross entropy loss between the predicted confidence vector $X$ and the ground truth vector $Y$ as in $\mathrm{\mathcal{L}_{anc}=BCE(X, Y)}$. The total loss used in the 3DJCG model would be $\mathcal{L}=\mathcal{L}_{org}+\mathcal{L}_{anc}$, where $\mathcal{L}_{org}$ represents the original losses used.

\section{Neural Speakers}
\subsection{SATCap-\datasetSuffix{}}
In \Cref{fig:sat_cap_model}, we show the SATCap-\datasetSuffix{} model, which is discussed in Section 4.2.1 in the main paper. The SATCap-\datasetSuffix{} model is based on the ``Show, Attend, and Tell" model, which is a 2D image captioning model. To make it amenable to purely 3D inputs, we replace the image encoder with the encoder network found in the MVT model~\cite{huang2022multi}, which is a point cloud PointNet++ encoder together with 3D object self-attention layers. For the decoder network, we use a unidirectional LSTM cell~\cite{lstm}. The speaker model is trained via teacher-forcing~\cite{teacher_forcing}. our proposed entity prediction loss $\mathcal{L}_{ent}$ is applied during the decoding steps in the following manner. At each decoding step, if the current word to be predicted corresponds to a scan entity (table and fridge words in \Cref{fig:sat_cap_model}), our  loss pushes the object(s) corresponding to the underlying scan entity to be the highest scoring among all objects present in the input scene. The entity prediction loss is not applied if the current word to be predicted does not correspond to a scan entity.

\begin{figure*}[!htb]
    \centering
    \includegraphics[width=0.6\linewidth]{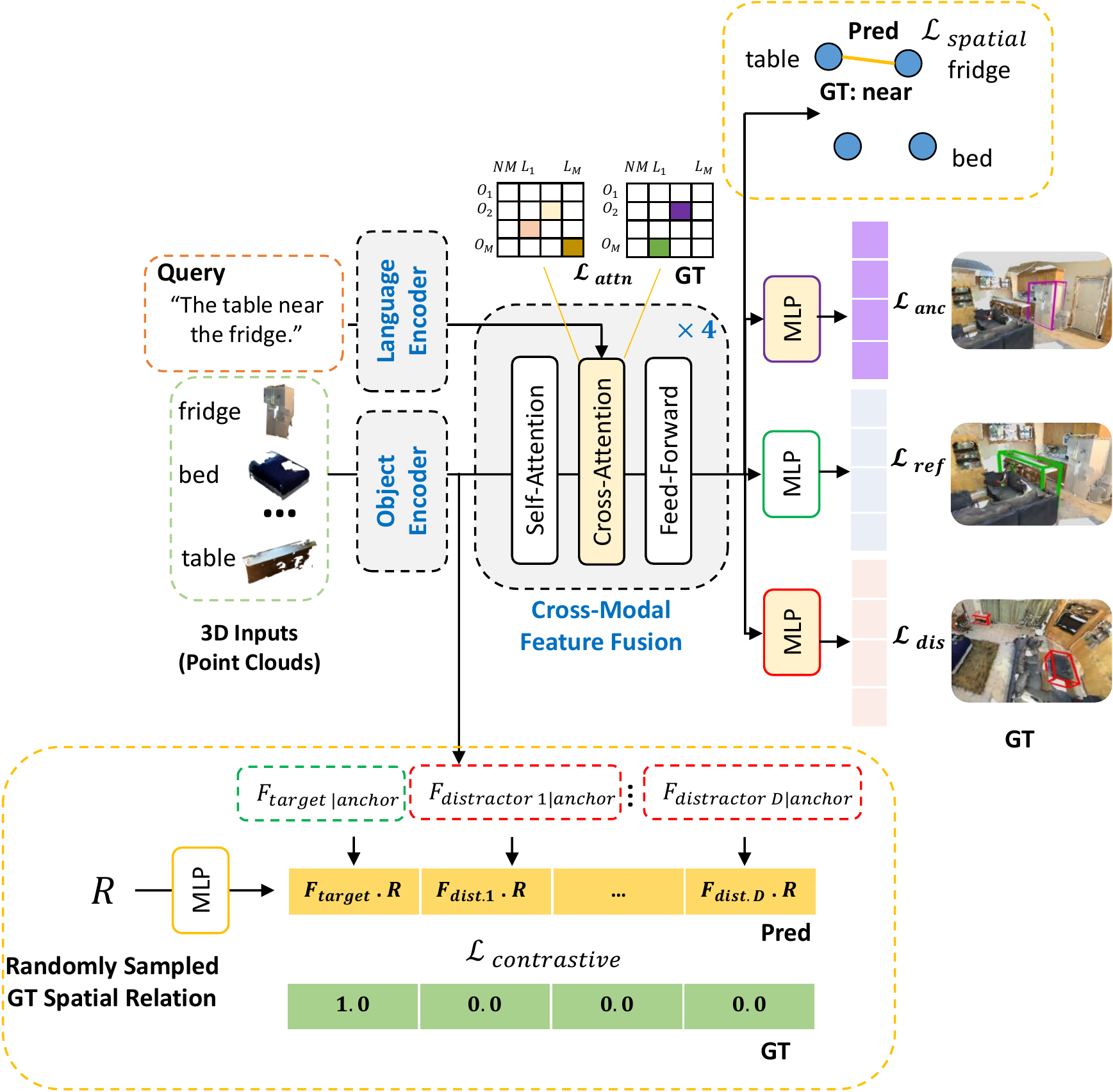}
    \caption{\textbf{Our proposed modifications to MVT-\datasetSuffix{} for exploiting the pair-wise spatial relationships that improve the  listening performance.} We propose two losses; $\mathcal{L}_{contrastive}$ and $\mathcal{L}_{spatial}$. $\mathcal{L}_{contrastive}$ aims at better understanding the spatial relationship between the target object and an anchor object while contrasting the spatial relation between the anchor and the same-class distractor objects. The $\mathcal{L}_{spatial}$ aims at predicting the spatial relationships between the object pairs where their ground truth spatial relationship is known.}
    \label{fig:spa_mvt}
\end{figure*}

\section{Implementation Details}
For the listening experiments, we used the same hyper-parameters specified in MVT~\cite{huang2022multi} and SAT~\cite{SAT_3D}. For the 3D object localization experiment, we use the same hyper-parameters of 3DJCG~\cite{cai20223djcg}. We use one NVidia V100 GPU in each of our experiments. We use the same hyper-parameters found in~\cite{Yang_2020_CVPR} for the neural speakers.

\section{Ablation Studies and More Quantitative Results}

\paragraph{Usefulness of exploiting the pairwise spatial relationships}
We exploit the extra annotations of the extracted pairwise spatial relationships discussed in Section 3.2 in the main paper. In \Cref{fig:spa_mvt}, we show our modifications to MVT-\datasetSuffix{} neural listener. We introduce two losses that exploit the pair-wise spatial relations. The first loss $\mathcal{L}_{contrastive}$ is a contrastive loss that operates as follows; for a training example, we randomly sample a ground-truth spatial relationship between the target object and an anchor object (the relationship does not necessarily present in the input utterance). The sampled spatial relationship is valid between the target object and the anchor while it is valid for none of the same-class distractor objects. We embed the spatial relation class into a vector $R$ with dimension $d$. We then concatenate the object feature (computed by the PointNet++ encoder~\cite{qi2017pointnetpp}) of the anchor object to the target object feature and the feature of each of the same-class distractor objects. The concatenated features are then transformed using an MLP and the generated features are called $F$ each of dimension $d$ as shown in \Cref{fig:spa_mvt}. We then apply a cosine similarity between the embedded feature of the spatial relation $R$ and each of the $F$ features. The   $\mathcal{L}_{contrastive}$ loss is the cross entropy between the predicted distribution and the ground-truth vector which is a one-hot vector, where the value of one corresponds to the target object. The second loss is called $\mathcal{L}_{spatial}$ and it operates on the context-aware features that are computed after the cross-modal fusion between the 3D objects and the input language and it works in the following manner. For each of the object pairs where the ground-truth spatial relationships are known, we apply a spatial relation classification loss on the concatenated features of the object pairs. To summarize, the spatial relationship losses are defined as $\mathcal{L}_{rel} = \mathcal{L}_{contrastive} + \mathcal{L}_{spatial}$.

As shown in \Cref{tab:spa_results}, we observe an improvement in the listening performance when combining the spatial relationship losses with both the anchor prediction loss and the same-class distractor loss. However, the performance didn't improve when using all four losses together.

\begin{table}[!htb]
    \centering
    \begin{tabular}{ccccc}
     \multicolumn{1}{c}{$\mathcal{L}_{attn}$} & \multicolumn{1}{c}{$\mathcal{L}_{anc}$}& \multicolumn{1}{c}{{$\mathcal{L}_{dis}$}} & 
     {{$\mathcal{L}_{rel}$}} & 
     \multicolumn{1}{c}{{Overall}}  \\
     \toprule

 \checkmark & \checkmark & &  &  
        58.7\%  \\ 
    \checkmark & \checkmark & \checkmark &  &  
    {59.3\%} 
    \\ 
    \checkmark & \checkmark & & \checkmark &  
        \textbf{59.7\%}  \\ 
        
    \checkmark & \checkmark & \checkmark & \checkmark &  
        {59.3\%} 
        \\ 
    \bottomrule
    \end{tabular}
    \caption{\textbf{Ablation study on using our proposed losses that exploit the extracted spatial relationships in the MVT-\datasetSuffix{} model.} Our proposed losses cause an improvement in the listening performance (+1.0\%) when being used with the anchor prediction loss and the same-class distractor loss.}
    \label{tab:spa_results}
\end{table}

\begin{table*}[th!]
\begin{myresizeenv}
\footnotesize
    \centering
   \begin{tabular}{ccccccccc}

    \\
    \multicolumn{1}{c}{$\mathcal{L}_{attn}$} & \multicolumn{1}{c}{$\mathcal{L}_{anc}$}& \multicolumn{1}{c}{{$\mathcal{L}_{dis}$}} & \multicolumn{1}{c}{{Overall}} & \multicolumn{1}{c}{Easy} & \multicolumn{1}{c}{Hard}& \multicolumn{1}{c}{View-dep.}&\multicolumn{1}{c}{View-indep.} & 
    \\
     \toprule
      & &  &  
        55.1\%$\pm$0.3\% & 
        61.3\%$\pm$0.4\% & 
        49.1\%$\pm$0.4\% & 
        54.3\%$\pm$0.5\% & 
        55.4\%$\pm$0.3\% &  
        \\ 
     \checkmark & & &  
        56.6\%$\pm$0.2\% &  
        63.0\%$\pm$0.3\% &  
        50.5\%$\pm$0.3\% &  
        55.4\%$\pm$0.4\% &  
        57.2\%$\pm$0.2\% &
        \\ 
     & & \checkmark &  
        56.9\%$\pm$0.3\% &  
        63.5\%$\pm$0.3\% &  
        50.6\%$\pm$0.3\% &  
        55.3\%$\pm$0.4\% &  
        57.8\%$\pm$0.4\% &
        \\ 

    \checkmark & & \checkmark &  
        57.4\%$\pm$0.3\% &  
        64.3\%$\pm$0.4\% &  
        50.8\%$\pm$0.4\% &  
        55.6\%$\pm$0.6\% &  
        58.3\%$\pm$0.3\% &
        \\ 
    & \checkmark & \checkmark &  
        57.9\%$\pm$0.2\% &  
        63.7\%$\pm$0.2\% &  
        52.3\%$\pm$0.2\% &  
        56.0\%$\pm$0.2\% &  
        58.9\%$\pm$0.3\% &
        \\ 
     & \checkmark & &  
        58.1\%$\pm$0.3\% &  
        63.8\%$\pm$0.5\% &  
        52.6\%$\pm$0.3\% &  
        56.7\%$\pm$0.3\% &  
        58.8\%$\pm$0.4\% &
        \\ 
    \checkmark & \checkmark & &  
        58.7\%$\pm$0.3\% &  
        64.6\%$\pm$0.4\% &  
        53.1\%$\pm$0.4\% &  
        \textbf{57.5\%$\pm$0.3\%} &  
        59.3\%$\pm$0.4\% &
        \\ 
    \checkmark & \checkmark & \checkmark &  
        \textbf{59.3\%$\pm$0.1\%} &  
        \textbf{65.4\%$\pm$0.3\%} &  
        \textbf{53.5\%$\pm$0.2\%} &  
        57.3\%$\pm$0.3\% &  
        \textbf{60.4\%$\pm$0.2\%} &
        \\ 
    \bottomrule
    \end{tabular}
    \caption{\textbf{Ablation study on neural listeners.} We ablate different combinations of our proposed auxiliary losses on the MVT neural listener, trained on \nr{} using our proposed \datasetName{} dataset.}
    \label{tab:ablationOnLosses2}
    \end{myresizeenv}
\end{table*}

\paragraph{Performance of listener with an increasing number of scan entities.} In \Cref{fig:comp_anchors}, we observe that the listening performance decreases when the difficulty of the input utterances increases where more scan entities and same-class distractor objects are involved. MVT-\datasetSuffix{} performs better than the original MVT model.

\paragraph{Effectiveness of the pre-trained encoder in the M2cap-\datasetSuffix{}.} In \Cref{tab:speaker_ablation}, we show the usefulness of using a pre-trained object encoder (trained on the neural listening task), which is discussed in Section 4.2.2 in the main paper. The usage of the pre-trained encoder improves the performance of the M2Cap-\datasetSuffix{} neural listener in all of the four captioning metrics on the \nr{} dataset.

\begin{table*}[!htb]
    \centering
    \begin{tabular}{lcccc}
\multicolumn{1}{c}{Arch.} & 
\multicolumn{4}{c}{\nr{}}\\
\toprule
   & C  &  B-4 &  M & R \\
\hline
M2Cap
&
86.15 &
37.03 &
30.63 &
67.00 \\
M2Cap-\datasetSuffix{} w/o Pre-trained Encoder &
{88.68} &
{37.29} &
{31.06} &
{67.35} \\
M2Cap-\datasetSuffix{} &
\textbf{93.25} &
\textbf{39.33} &
\textbf{31.55} &
\textbf{68.33} \\
\bottomrule
\end{tabular}
    \caption{\textbf{Effectiveness of using the pre-trained object encoder in M2Cap-\datasetSuffix{}}. Using the pre-trained object encoder helps improve the performance of M2-Cap-\datasetSuffix{} neural speakers in all of the four metrics.}
    \label{tab:speaker_ablation}
\end{table*}

\paragraph{Effectiveness of losses in MVT-\datasetSuffix{}.}  In \Cref{tab:ablationOnLosses2}, we show an ablation study upon using our proposed losses on the MVT-\datasetSuffix{} neural listeners. Following~\cite{achlioptas2020referit_3d}, we do testing using five random seeds, and we report the mean and the standard deviation of the accuracy.

% \section{More Qualitative Results}
% In \Cref{fig:qual_listern}, we show more qualitative results for our proposed model MVT-\datasetSuffix{}. 

% \begin{figure*}[!htb]
%     \centering
%     \includegraphics[width=\linewidth]{Figures/listen_qual-cropped.pdf}
%     \caption{\textbf{More qualitative results of MVT-\datasetSuffix{} compared to the original MVT model.}}
%     \label{fig:qual_listern}
% \end{figure*}

\begin{figure*}[!htb]
    \centering
    \includegraphics[width=0.8\linewidth]{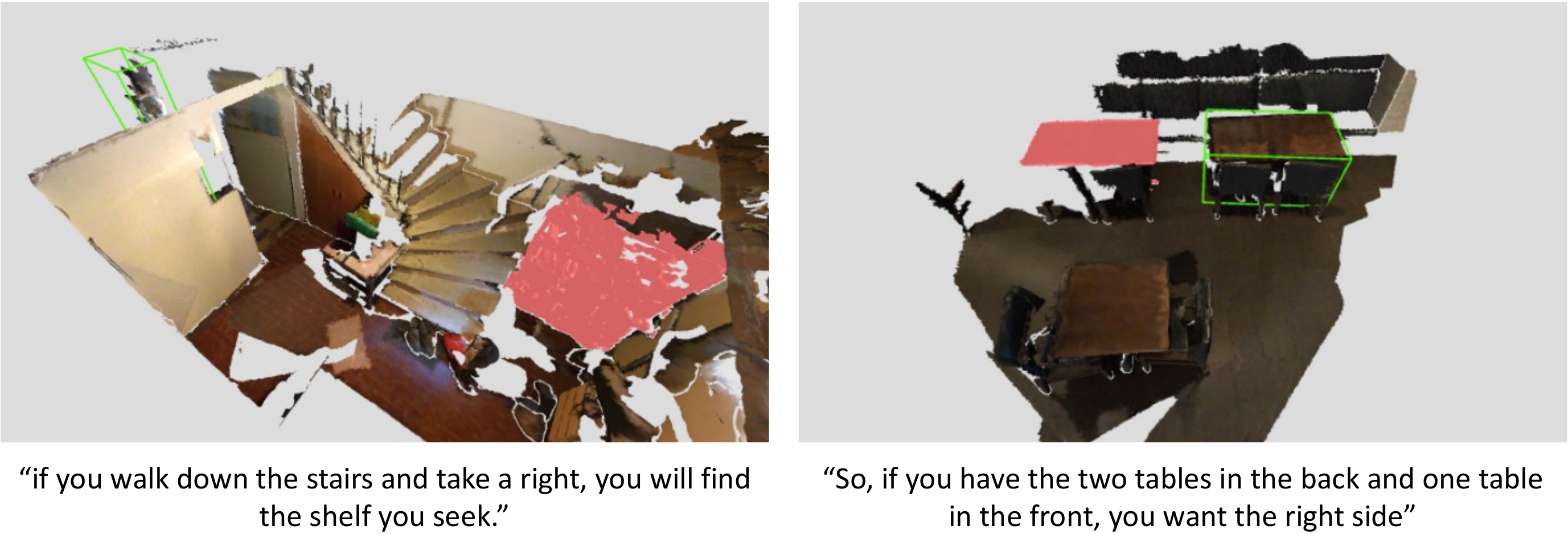}
    \caption{\textbf{Failure examples where the MVT-\datasetSuffix{} model struggles to identify the target object (green) because of the complex language descriptions. The incorrect predictions are highlighted in red color.}}
    \label{fig:failureExamples}
\end{figure*}

\section{Limitations}
Our extension of Nr3D and ScanRefer with \datasetName{} is based on the original utterances in these two datasets. Hence, we are constrained in a linguistic corpus where the grounding language used is English. It would be of interest to explore the efficacy of our method and annotation approach to other languages, especially to reduce the possible biases a restrictive set of cultural groups might be introducing. Moreover, despite achieving SoTA results in two popular and essential tasks for 3D-based visio-linguistic grounding tasks, it is clear that our methods are not yet on par with human-level performance (see \cref{fig:failureExamples}). More studies around competing methods, the underlying supervision used, and even transfer-learning approaches that can leverage e.g., large-scale 2D-based data, or recent foundational models, are expected to be fruitful in closing the gap between learning-based methods and human efficacy.

\end{document}